\documentclass[letterpaper]{article} 
\usepackage{aaai2026}  
\usepackage{times}  
\usepackage{helvet}  
\usepackage{courier}  
\usepackage[hyphens]{url}  
\usepackage{graphicx} 
\usepackage{natbib}  
\usepackage{caption} 
\usepackage{algorithm}
\usepackage{algorithmic}
\usepackage{color}
\usepackage{pifont}
\usepackage{multirow}
\usepackage{subfigure}
\usepackage{amsmath}
\usepackage{amssymb}
\usepackage{amsthm}

\newtheorem{lemma}{Lemma}
\newtheorem{proposition}{Proposition}
\newtheorem{corollary}{Corollary}

\usepackage{newfloat}
\usepackage{listings}
\DeclareCaptionStyle{ruled}{labelfont=normalfont,labelsep=colon,strut=off} 
\floatstyle{ruled}
\newfloat{listing}{tb}{lst}{}
\floatname{listing}{Listing}
\title{FedShard: Federated Unlearning with Efficiency Fairness and Performance Fairness}
\author{
    Siyuan WEN\textsuperscript{\rm 1}, 
    Meng ZHANG\textsuperscript{\rm 2}, 
    Yang YANG\textsuperscript{\rm 3}, 
    Ningning DING\thanks{Corresponding author.}\textsuperscript{\rm 1}
}
\affiliations{
    \textsuperscript{\rm 1}Hong Kong University of Science and Technology (Guangzhou)\\
    \textsuperscript{\rm 2}Zhejiang University\\
    \textsuperscript{\rm 3}Hong Kong University of Science and Technology\\
    
    swen211@connect.hkust-gz.edu.cn, mengzhang@intl.zju.edu.cn, yangyangsh@ust.hk, ningningding@hkust-gz.edu.cn}

\usepackage{bibentry}

\begin{document}
\newcommand\tcl[1]{#1}
\newcommand\tdl[1]{#1}
\newcommand\sftl[1]{\textbf{#1}}
\newcommand\sfttcl[1]{\textbf{#1}}
\newcommand\sfttdl[1]{\textbf{#1}}

\newcommand\ftl[1]{\textbf{#1}}
\newcommand\fttcl[1]{\textbf{#1}}
\newcommand\fttdl[1]{\textbf{#1}}

\newcommand\fbl[1]{\underline{#1}}
\newcommand\fbtcl[1]{\underline{#1}}
\newcommand\fbtdl[1]{\underline{#1}}   

\maketitle

\begin{abstract}
    To protect clients' right to be forgotten in federated learning, federated unlearning aims to remove the data contribution of leaving clients from the global learned model. While current studies mainly focused on enhancing unlearning efficiency and effectiveness, the crucial aspects of efficiency fairness and performance fairness among decentralized clients during unlearning have remained largely unexplored. In this study, we introduce FedShard, the first federated unlearning algorithm designed to concurrently guarantee both efficiency fairness and performance fairness. FedShard adaptively addresses the challenges introduced by dilemmas among convergence, unlearning efficiency, and unlearning fairness. Furthermore, we propose two novel metrics to quantitatively assess the fairness of unlearning algorithms, which we prove to satisfy well-known properties in other existing fairness measurements. Our theoretical analysis and numerical evaluation validate FedShard's fairness in terms of both unlearning performance and efficiency. We demonstrate that FedShard mitigates unfairness risks such as cascaded leaving and poisoning attacks and realizes more balanced unlearning costs among clients. Experimental results indicate that FedShard accelerates the data unlearning process 1.3-6.2 times faster than retraining from scratch and 4.9 times faster than the state-of-the-art exact unlearning methods.
\end{abstract}

\section{Introduction}
Federated learning (FL) is a prominent paradigm for collaborative machine learning on distributed data, 
wherein decentralized clients train a shared global model under the coordination of a central server~\cite{fedavg}. 
To protect the ``right to be forgotten" (RTBF) for clients who leave the FL system, federated unlearning (FU) has emerged as an essential method for removing the contributions of leaving data from the global model~\cite{GDPR,CCPA,privacyconcerns,FLsurvey}. 
However, prior work in FU mainly focuses on the efficiency and effectiveness of the unlearning process~\cite{mulsurvey,mulsurvey2}. We argue that the fairness of federated unlearning—with respect to both computational costs and performance impacts—is a critical yet overlooked aspect, especially for system sustainability and security.

\begin{figure}[t]
    \centering
    \includegraphics[width=0.4\textwidth]{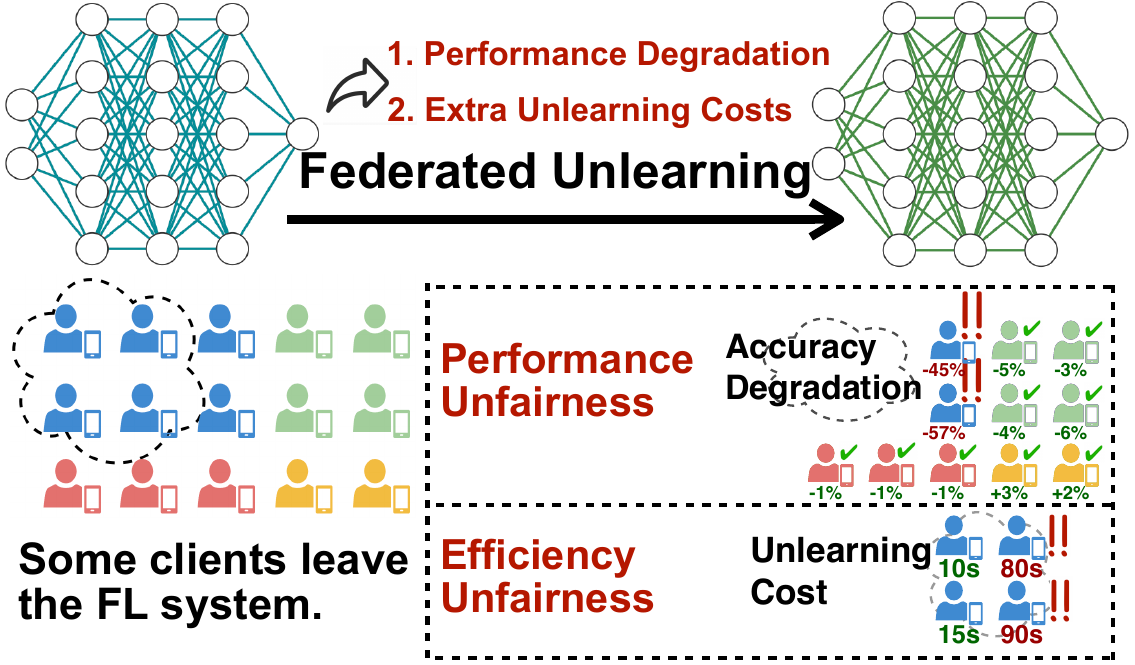}
    \caption{An illustration of performance unfairness and efficiency unfairness in federated unlearning. Red exclamation marks indicate that clients encounter unfairness.}~\label{fig:arch}
\end{figure}

As illustrated in Figure \ref{fig:arch}, literature usually overlooked two types of issues, performance fairness (\textbf{P.F.}) and efficiency fairness (\textbf{E.F.}), which are crucial and practically important. 
P.F. aims to achieve a fair performance degradation caused by unlearning among clients, guaranteeing both remaining clients' rights to good model performance, and leaving clients' RTBF. Without P.F., unlearning leaving clients may cause an excessive accuracy decrease on specific data, thereby encouraging other remaining clients (especially those with similar data) to leave. Such cascaded leaving can severely compromise the sustainability of an FL system. Moreover, unfair unlearning can be further exploited for data poisoning attacks~\cite{dpa1}. By deliberate unlearning requests, malicious clients can launch data poisoning attacks on specific clients with similar data and compromise the security of the FL system. E.F. demands a fair unlearning cost for every client when it leaves. Unlearning methods without E.F. may harm the system sustainability by discouraging clients from participating in the system, as they are concerned about unfairly large leaving costs.

\begin{figure}[t]
        \centering
        \subfigure[]{
        \begin{minipage}[t]{0.44\linewidth}
            \centering
            \includegraphics[width=\textwidth]{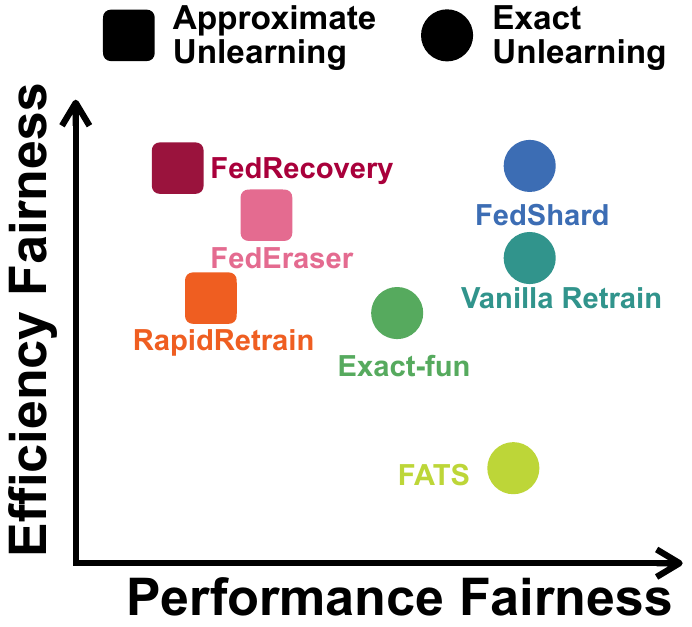}
        \end{minipage}%
    }
        \subfigure[]{
        \begin{minipage}[t]{0.48\linewidth}
            \centering
            \includegraphics[width=\textwidth]{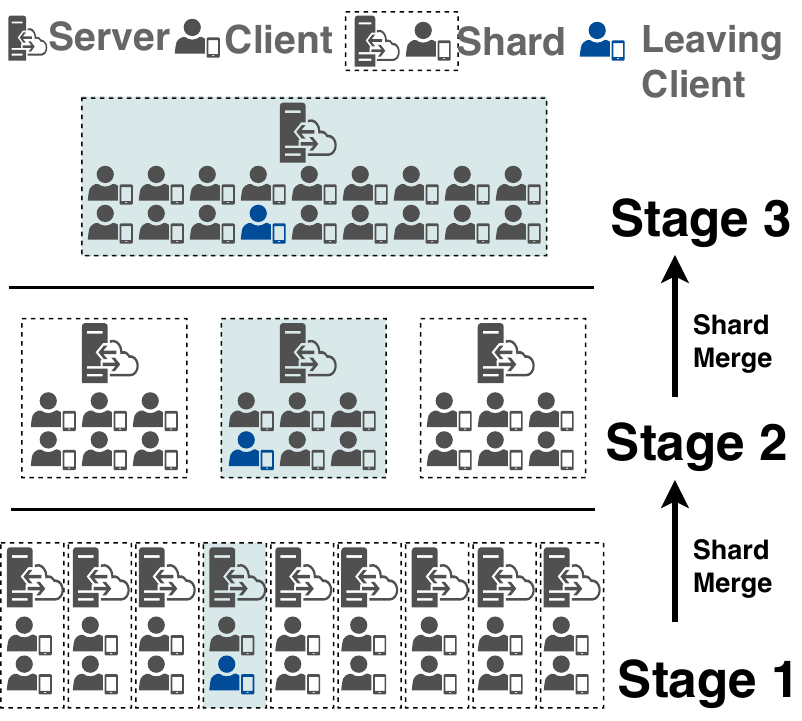}
        \end{minipage}%
    }
        \caption{Left: A conceptual illustration showing that only FedShard simultaneously achieves both performance fairness and efficiency fairness. Right: An example of the FedShard architecture with a merging rate of 3. Clients are organized in a tree-like structure. When clients request unlearning (indicated in blue), only the shards on the direct path to the root (dark background) require retraining.}~\label{fig:arch1}
\end{figure}

Current FU methods are unsatisfactory in terms of performance fairness and efficiency fairness. Specifically, related work can be classified into two categories: calibration-based approximate unlearning~\cite{FedAF,ShardingE,verifi,FYEFUL} and retraining-based exact unlearning~\cite{exactfun}. As illustrated in Figure \ref{fig:arch1}(a), calibration unlearning (such as FedRecover~\cite{fedrec}, FedEraser~\cite{federaser} and RapidRetrain~\cite{RR}) often introduces extra performance degradation on remaining clients or insufficient unlearning on leaving clients, which compromises the P.F. Exact retraining (such as FATS~\cite{FATS}) can easily introduce significant variance in unlearning efficiency, thereby violating E.F..

To address these shortcomings, we propose FedShard, a novel sharded federated unlearning framework that prioritizes both fairness and efficiency to achieve effective federated unlearning. As illustrated in Figure \ref{fig:arch1}(b), the FedShard framework operates in two repeating phases: isolated federated training within distinct client shards, and a progressive merging of these shards until all clients are integrated into a single shard. This ``isolated yet integrate'' design inherently minimizes the impact of a client's leaving and facilitates highly efficient unlearning by leveraging cached intermediate model parameters.

However, designing such a sharded framework presents non-trivial challenges among convergence, unlearning efficiency and unlearning fairness. First, determining appropriate stage-wise training rounds is critical. 
Excessive training within a shard can cause its model to overfit to local data and diverge from the global optimum, whereas insufficient training results in poor convergence.
Second, it presents another challenge to determine which clients should be merged into which new shards in the next stage. While a strategy that minimizes training time disparities across clients would promote efficiency fairness, it may compromise the convergence speed and final performance of the global model. To navigate these trade-offs, we develop two adaptive algorithms that dynamically determine the appropriate training rounds and shard merging configurations, ensuring both fairness and overall efficiency.

Our primary contributions are summarized as follows:
\begin{itemize}
    \item \textit{Novel sharded framework with adaptive algorithms.} We introduce shards into federated learning and unlearning, providing a novel framework that is both isolated and integrated. We further develop adaptive algorithms to tackle the challenges of stage-wise training rounds and shard merging strategy for sharded training. 
    \item \textit{Fair and efficient federated unlearning.} We propose the first efficient federated unlearning framework that satisfies both performance fairness and efficiency fairness. We also prove that FedShard unlearns at least $P/{2}$ times faster than retraining from scratch, where $P$ is the number of stages. Furthermore, FedShard does not restrict to a single unlearning request processing but allows multiple simultaneous unlearning requests, which provides additional efficiency improvement.
    \item \textit{Fairness metrics for federated unlearning.} We identify the performance fairness and efficiency fairness requirements for federated unlearning and introduce novel metrics to measure them for different FU methods. We further validate the two metrics through theoretical analysis and numerical evaluation, showing that they are practical and effective.
    \item \textit{Evaluation contribution.} Our evaluation demonstrates the fairness weakness of existing state-of-the-art works and confirms FedShard's ability to address fairness issues, including cascaded leaving, data poisoning attacks, and unfair-distributed unlearning costs. Our experiment results also show that FedShard unlearns 1.3-6.2 times faster than retraining and up to 4.9 times faster than the state-of-the-art exact unlearning methods.
\end{itemize}

\section{Related Work}
Related research of federated unlearning can be categorized into two primary approaches: calibration-based approximate unlearning and exact retraining.\footnote{Further literature related to FL fairness and shard methods can be found in Appendix A due to space limit.}

Parameter calibration methods scale down leaving clients' gradients with the Hessian matrix~\cite{RR}, cached intermediate information~\cite{federaser}, or gradient ascent~\cite{FYEFUL}. They achieve efficient federated unlearning, while such methods can easily cause extra performance degradation~\cite{fedrec} on remaining clients and insufficient unlearning on leaving clients, violating P.F. Our proposed method will avoid such issues and achieve performance fairness while maintaining efficiency.

Exact retraining methods achieve quite effective unlearning, \textit{e.g.}, FATS~\cite{FATS} and Exact-fun~\cite{exactfun}. They utilize cached information to alleviate the heavy overhead caused by vanilla retraining. Unfortunately, these methods usually lead to disparities in unlearning costs, compromising efficiency fairness. For example, \citeauthor{FATS} allow clients to join at different rounds for fast federated unlearning. When the later-joined client is leaving, earlier rounds without its participation can be bypassed from unlearning, thereby achieving efficient unlearning. However, earlier-joined clients will burden extra overheads, leading to unfair unlearning costs. Our FedShard ensures that every client has a closer unlearning cost related only to its own data contribution, thereby ensuring a fair unlearning cost while maintaining efficient retraining.

\section{FedShard Design}~\label{sec:fedshard}
In this section, we show the architecture and mechanics of FedShard in detail. We first describe the sharded federated learning and unlearning process in \S\ref{sec:sfl}. We then present the two core adaptive algorithms that enable the fairness and efficiency: a cluster-based shard merging strategy (\S\ref{sec:a1}) and a variance-aware training round allocation scheme (\S\ref{sec:a2}).

\subsection{Sharded Federated Learning and Unlearning}\label{sec:sfl}
\begin{figure}
    \centering
    \includegraphics[width=0.42\textwidth]{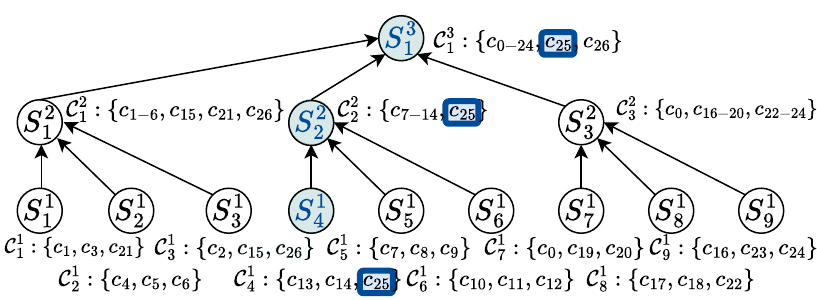}
    \caption{Illustration of the FedShard learning and unlearning process of the example in Figure~\ref{fig:arch1}. Each node represents a shard, and each layer presents a training stage. Arrows indicate the merging of shards between stages. When client $c_{25}$ requests unlearning, only the shards on its hierarchical path (dark background) require retraining.}~\label{fig:arch2}
    \vspace{-10pt}
\end{figure}

FedShard efficiently achieves exact unlearning by organizing clients into a hierarchical and multi-stage structure. As illustrated in Figure~\ref{fig:arch2}, this structure can be conceptualized as a tree, where clients are the leaves, shards are the nodes, and the training stages represent the levels of the tree. For the convenience of understanding, we formalize the items in FedShard with the following notations. We provide the complete symbol list in Appendix B for quick viewing.

Let $\mathcal{C}=\{c_1, \dots, c_K\}$ be the set of all clients. An entire FedShard process consists of $P\triangleq \lceil K/R \rceil$ stages, where $K$ is the total number of clients and $R$ is a user-defined merging rate to decide how many shards are merged into one super-shard at the next stage. At each stage $p \in \{1, \dots, P\}$, there are $N_p$ shards, denoted by $\mathcal{S}^p = \{S_1^p, \dots, S_{N_p}^p\}$. Each shard $S_s^p$ independently performs federated learning and unlearning in parallel, and we denote its model parameters as $\theta_s^p$. Specially, $\theta_s^{p,0}$ refers to the initial parameters for shard $S_s^p$. The aggregation weight for $S_s^p$ is denoted as $w_s^p$. We define the contribution factor $\alpha_c$, a key concept to measure a client's impact on the global model, as the angle between the local update vector $\Delta\theta_c$ and the global update vector $\Delta\theta$:\footnote{Our framework is also applicable to other metrics for measuring the contribution factor, such as the loss function value, shapley value, etc. For the convenience of understanding, we use parameter similarity, the most intuitive one, in the main paper and provide discussion on other contribution factors in Appendix C.}
\begin{equation}~\label{eq:dis}
    \alpha_c=\arccos \frac{\left\langle\Delta \theta, \Delta \theta_c\right\rangle}{\|\Delta \theta\|_2\|\Delta \theta_c\|_2}.
\end{equation}
A smaller angle $\alpha_c$ implies a larger contribution to the global model. We use $\alpha_s^p$ to denote the average of $\alpha_c$s in $S_s^p$. 

The FedShard training workflow proceeds stage-by-stage. Each stage $p$ involves the following four phases: (1) \textit{shard generation}, (2) \textit{shard initialization}, (3) \textit{parallel federated training}, and (4) \textit{state caching}. For the first two phases, the server merges shards from stage $p-1$ to generate new shards using algorithm $\mathcal{A}_1$ (introduced later in \S\ref{sec:a1}). Each new shard is initialized and the number of training rounds for all shards is determined by algorithm $\mathcal{A}_2$ (introduced later in \S\ref{sec:a2}). For the last two phases, the server performs federated training within new shards and caches the model updates. We provide the detailed workflow, pseudocode and space complexity analysis in Appendix D.1 due to the space limits. This process repeats, with the number of shards decreasing at each stage, until a single global shard containing all clients is formed in the final stage $P$. 

This isolated and hierarchical training process is the key to efficient unlearning. When a client requests to be forgotten, retraining is only required for the shards that lie on its direct path to the root of the tree. All other shards can reuse their cached model states, significantly reducing the computational cost compared to retraining from scratch. 
For example, as shown in Figure~\ref{fig:arch2}, if client $c_{25}$ requests unlearning, only shards $S_4^1$, $S_2^2$, and $S_1^3$ need to be retrained. The unlearning procedure is detailed in Algorithm~\ref{alg:a3}, and its efficiency will be formally analyzed in \S\ref{sec:analysis}.

\begin{algorithm}[ht]
    \caption{FedShard Federated Unlearning Algorithm}
    \label{alg:a3}
    \textbf{Input}: Leaving client $c_l$, cached information during federated training FLCache

    \textbf{Output}: Unlearned model $\theta^*_u$

    \begin{algorithmic}[1]
        
        \STATE AShards = $\emptyset$ //\textit{ Affected shards}
        \STATE P $\leftarrow$ FLCache


        \FOR{$p$ {\bfseries =} 1, 2, $\dots$, P}

            \STATE S$_1^p$, S$_2^p$, $\dots$, S$_{N_p}^p$ $\leftarrow$ FLCache

            \STATE {\bfseries for} {$s$ {\bfseries =} 1, 2, $\dots$, N$_p$}, AShards.append(S$^p_s$) {\bfseries if} $c_l$ {\bfseries in} S$_s^p$
        \ENDFOR
        \FOR{$p$ {\bfseries =} 1, 2, $\dots$, P}
            \STATE\textit{// Phase 1\&2: Load shard S$_s^p$, its training round T$_s^p$, and its initial model $\theta_s^{p,0}$ from FLCache}

            S$_s^p$, T$_s^p$, $\theta_s^{p,0}$ $\leftarrow$ FLCache, $s$ {\bfseries =} 1, 2, $\dots$, N$_p$
            
            \STATE\textit{// Phase 3: Parallel federated training }
            \FOR{each $S_s^p$ {\bfseries in} stage $p$, s{\bfseries =} 1, 2, $\dots$, N$_p$}
                \IF{$S_s^p$ {\bfseries in} AShards}
                    \STATE$\theta_s^p$={\bfseries FederatedTraining}($\theta_s^{p,0}$, [clients$\in S_s^p$], $T_s^p$)
                \ELSE
                    \STATE$\theta_s^p$ $\leftarrow$ FLCache
                \ENDIF
            \ENDFOR
            \STATE\textit{// Phase 4: Cache the retrained shards}
            \STATE FLCache $\leftarrow$ S$_s^p$, $\theta_s^p$, $\alpha_s^p$, $w_s^p$ {\bfseries if} $S_s^p$ {\bfseries in} AShards
        \ENDFOR
    \end{algorithmic}
    \textbf{Return}: $\theta^*_u=\theta_1^P$ //\textit{Final model from the last stage}
\end{algorithm}

\subsection{Shard Merging Algorithm $\mathcal{A}_1$}\label{sec:a1}

\begin{figure}[ht]
    \centering
    \includegraphics[width=0.3\textwidth]{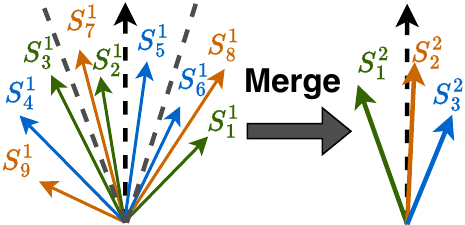}
    \caption{Example of the shard merging strategy. To form a new shard for the next stage, FedShard combines shards with diverse update directions (represented by arrows) to promote stable convergence.}~\label{fig:merge}
\end{figure}
The strategy for merging shards between stages is critical yet challenging. A naive approach presents a dilemma: merging shards with divergent model update directions (\textit{e.g.}, $S_9^1, S_6^1, S_1^1$ in Figure \ref{fig:merge}) can hinder convergence\footnote{As the experiment we later show in \S\ref{sec:evaluation}.}, while merging shards with highly similar update directions (\textit{e.g.}, $S_9^1, S_4^1, S_3^1$ in Figure \ref{fig:merge}) can create an unrepresentative super-shard that biases the global model. This bias would lead to inequitable unlearning costs, thereby compromising efficiency fairness.

Our merging algorithm $\mathcal{A}_1$ navigates this trade-off by promoting directional diversity within newly formed shards. The core idea is to first cluster the existing shards based on their model update directions (relative to the global average) and then to form new shards by selecting constituents from different clusters. 
For instance, in Figure \ref{fig:merge}, the new shard $S_2^2$ is intentionally formed by merging $S_7^1$ (with an average update direction), $S_8^1$ (with a large positive angle), and $S_9^1$ (with a large negative angle). This balanced approach ensures each new shard to be representative of the broader client population, leading to more stable convergence and a fairer distribution of learning contributions. Further details of the algorithm are provided in Appendix D.2.

\subsection{Training Round Allocation Algorithm $\mathcal{A}_2$}\label{sec:a2}
Determining the number of training rounds per shard at each stage is another critical challenge. Too many rounds can cause a shard's model to overfit on its local data, making it difficult to merge effectively and wasting computation. Conversely, too few rounds can lead to under-training, shifting an excessive convergence burden to the final stage, which in turn increases the unlearning cost for all clients.

Our allocation algorithm $\mathcal{A}_2$ addresses this by dynamically assigning training rounds based on the heterogeneity of clients within each shard. The intuition is that shards with more homogeneous clients (i.e., lower contribution variance) can be trained for more rounds without overfitting, while heterogeneous shards require fewer rounds to prevent their models from diverging.
Specifically, after determining a valid range of training rounds $[T^*_0, T^*_1]$, the number of rounds $T_s^p$ for shard $s$ in stage $p$ is set using the following inverse relationship:
\begin{equation}~\label{sec:adatra}
    \frac{T_s^p-T^*_0}{T^*_1-T^*_0}=\left[\frac{\sigma^2(\alpha_s^p)-\sigma^2(\alpha)_{min}}{\sigma^2(\alpha)_{max}-\sigma^2(\alpha)_{min}}\right]^{-1}, 
\end{equation}
where $\sigma^2(\alpha_s^p)$ is the variance of client contributions within shard $s$, and $\sigma^2(\alpha)_{min/max}$ are the minimum/maximum variances across all shards in the stage. This policy promotes E.F. by avoiding extreme numbers of training rounds. More details of $\mathcal{A}_2$ is available in Appendix D.3.

\section{Analysis}\label{sec:analysis}
In this section, we provide a theoretical foundation for FedShard. We first prove that FedShard achieves exact unlearning, equivalent to retraining from scratch (\S\ref{sec:theory_effectiveness}). We then analyze its computational efficiency for both single-client and multi-client unlearning requests, demonstrating significant speedups over baseline methods (\S\ref{sec:theory_efficiency} and \S\ref{sec:mulul}).\footnote{All proofs are provided in Appendix E due to space limits.}

\subsection{Effectiveness of Unlearning}\label{sec:theory_effectiveness}


We hold the following proposition that is widely accepted and leveraged in the exact unlearning literature~\cite{SISA,FATS,exactfun}:
\begin{proposition}~\label{prop:1}
    For any training process $\theta^*=\mathcal{A}_{\text{fl}}({D}_0;\theta^0)$ with training dataset ${D}_0$ and initial model $\theta^0$ and $\forall {D}_l \in \mathcal{D}$, $$\text{if }({D}_l\not\subseteq{D}_0) \land (\theta^0 \text{ not affected by }{D}_l),$$ then we say $\theta^*$ is not affected by ${D}_l$ and using cached updates $\theta^*$ for unlearning ${D}_l$ is equivalent to retraining from scratch on ${D}_0\backslash{D}_l$.
\end{proposition}
With this proposition, we prove by induction to show that FedShard unlearning, which only trains the one shard containing the leaving client every stage, is as effective as retraining from scratch. 

\subsection{Efficiency of Single-Client Unlearning}\label{sec:theory_efficiency}

To build intuition, we first consider a simplified case with a constant number of training rounds ($T_0$) for all shards.
\begin{proposition}~\label{lemma:r1}
    FedShard FU is $r_1$ times faster compared with retraining from scratch, where
    \begin{equation}~\label{thrm:r1}
        r_1=\frac{R-1}{R}\frac{K}{K-1}{P}.
    \end{equation}
\end{proposition}

With $R\in\mathbb{Z}^+$, this proposition shows that FedShard unlearns at least ${P}/{2}$ $\times$ (usually ${P}\times$) faster than retraining the model from scratch through FedAvg, \textit{i.e.}, vanilla retraining. 

Next, we consider the general case where training rounds $T_s^p$ can vary across shards and stages, as determined by our algorithm $\mathcal{A}_2$. Let $T_{max}$ and $T_{min}$ be the maximum and minimum rounds assigned to any shard.

\begin{proposition}[Efficiency in the General Case]\label{prop:r1p}
In the general case with variable training rounds, the speedup factor $\hat{r_1}$ over vanilla retraining is bounded as follows:
\begin{equation}\label{thrm:r1p}
r_1 \frac{T_0}{T_{max}} \le \hat{r_1} \le r_1 \frac{T_0}{T_{min}}.
\end{equation}
\end{proposition}
This proposition reveals that the bound of training rounds ($T_{max}$ and $T_{min}$) directly impacts efficiency. It provides the theoretical justification to design Algorithm $\mathcal{A}_2$, which seeks to maintain a narrow bound for high efficiency and fairness.

\subsection{Multiple Unlearning and Time Complexity}\label{sec:mulul}
Next we show that unlearning multiple clients together is more efficient in FedShard. Considering $m$ clients that leave at the same time, we denote $p'$ as the stage after which all shards contain leaving clients. Then we have: 
\begin{proposition}~\label{lemma:r2}
    Compared with unlearning one by one, performing multiple unlearning requests in FedShard achieves at least $r_2^-$ times and at most $r_2^+$ times faster, where:
    \begin{equation}\label{eq:r21}
        r_2^+=m, \quad
        r_2^-=\frac{R}{R-1}\frac{K-1}{K}\frac{m}{(P-p')}\ge 1.
    \end{equation} 
\end{proposition}
This proposition shows that multiple unlearning in FedShard is even more efficient compared to FedShard unlearning one by one. Based on Propositions \ref{lemma:r1}, \ref{prop:r1p} and \ref{lemma:r2}, we have:
\begin{corollary}~\label{cor:1}
    FedShard unlearning requires $\mathcal{O}(K)$ time for one client, and $\mathcal{O}(K\lg m)$ time for $m$ clients,
    where $K$ is the number of clients.
\end{corollary}

\section{Fairness Metrics}\label{sec:metr}

To quantitatively evaluate the fairness of FU algorithms, we introduce novel metrics for P.F. (\S\ref{sec:metrFp}) and E.F. (\S\ref{sec:metrFe}). We begin by additionally formalizing necessary notations.

Following an unlearning request, the set of clients $\mathcal{C}$ is partitioned into leaving clients $\mathcal{C}_L$ and remaining clients $\mathcal{C}_R$. We use $c_l\in\mathcal{C}_L$ and $c_r\in\mathcal{C}_R$ to specify a leaving client and a remaining client, respectively, \textit{e.g.}, $c_{l1}$, $c_{l2}$, and $c_{r1}$. We denote the local dataset of client $c$ as $D_c$, and denote the model's accuracy on a test set $D_c$ is given by $\mathcal{Y}(D_c)$. Based on $\alpha_c$ defined in Equation (\ref{eq:dis}),
${\rm Dis}(c_1, c_2) \triangleq |\alpha_{c_1} - \alpha_{c_2}|$ can represent the data distance between any two clients $\alpha_{c_1}$ and $\alpha_{c_2}$, where a smaller value indicates greater similarity in their data's impact on the model. 

\subsection{Performance Fairness Metric $M_p$}\label{sec:metrFp}
We define $M_p$ as the P.F. metric for federated unlearning:
\begin{equation}~\label{eq:mp}
        M_p \triangleq \frac{1}{|\mathcal{C}|}\sum_{c\in\mathcal{C}} f_{\oplus}(\Delta \mathcal{Y}({D}_c),\min_{c_r\in\mathcal{C}_R}{\rm Dis}(c,c_r)).
\end{equation}
$M_p$ captures whether the clients' local performance degradation caused by unlearning ($\Delta \mathcal{Y}({D_c})$) is unfair considering clients' data uniqueness. We assess the uniqueness by $\min_{c_r\in\mathcal{C}_R}{\rm Dis}(c,c_r)$, the minimal data distance between the leaving client $c$ and any client $c_r$ remaining in the system. Specifically, $f_{\oplus}(x,y)=(\tilde{x}+\tilde{y})\cdot({\tilde{x}}^{-1}+\tilde{y}^{-1})$ is a convex function and reaches its minimum at $\tilde{x}=\tilde{y}$. Operation $\tilde{x}$ normalizes $x$ to (0, 1], \textit{i.e.}, $\tilde{x}$$=$$\epsilon+(x-x_{min})/(x_{max}-x_{min})$, where $\epsilon$ is a small constant to avoid division by zero.
A lower $M_p$ score indicates superior performance fairness.

We give two examples to illustrate the rationales behind $M_p$. First, suppose a leaving client $c_{l1}$ that has unique data in the system (\textit{i.e.}, $\min_{c_r\in\mathcal{C}_R}{\rm Dis}(c_r,c_{l1})$ is large). A fair FU algorithm will effectively forget data of client $c_{l1}$ (\textit{i.e.}, $\Delta\mathcal{Y}({D}_{c_{l1}})$ is also large) and receive a lower $M_p$ score, indicating being performance-fair.
Second, suppose that leaving client $c_{l2}$ and a remaining client $c_{r1}$ have the same data (\textit{i.e.}, $\min_{c_r\in\mathcal{C}_R}{\rm Dis}(c_r,c_{l2})={\rm Dis}(c_{r1},c_{l2})$ is small). It's unfair for client $c_{r1}$ if the unlearned model has an excessively large performance drop $\Delta\mathcal{Y}({D}_{c_{r1}})$ on its dataset, as client $c_{r1}$'s data is still in the system. Therefore, such an unlearning algorithm receives a high $M_p$ score, indicating poor P.F.. As we shall see in \S\ref{sec:evaluation}, our experiments validate the effectiveness of our proposed metrics in measuring fairness.

\subsection{Efficiency Fairness Metric $M_e$}\label{sec:metrFe}
We define the E.F. metric $M_e$ as the contribution-weighted variance of unlearning costs:
\begin{equation}~\label{eq:me}
    M_e\triangleq \frac{1}{|\mathcal{C}|}\sum_{c\in\mathcal{C}} \frac{(Z_{avg}-Z_c)^2}{|\alpha_c|}.
\end{equation}
E.F. captures whether clients' unlearning cost (\textit{i.e.}, computing time) $Z_c$ is aligned with their impact (\textit{i.e.}, contribution $|\alpha_c|$) to the global model. 
Specifically, the numerator represents the variance of federated unlearning costs between one client and the average $Z_{avg}$. The denominator $|\alpha_c|$ normalizes the cost by the client's impact. A lower $M_e$ score indicates superior efficiency fairness.

We also give an example to illustrate the rationales behind $M_e$. Suppose that there are two clients $c_1$ and $c_2$ in the system who made similar contributions to the global model (i.e., $\alpha_{c_1} \approx \alpha_{c_2}$). An unfair FU algorithm requires vastly different costs to unlearn them ($Z_{c_1} \gg Z_{c_2}$), resulting in a large cost variance and thus a very high $M_e$ score.

\subsection{Fairness Properties}\label{sec:metrP}
It is important to note that the absolute values of $M_p$ and $M_e$ are only meaningful when comparing different FU methods under identical settings. Furthermore, $M_p$ and $M_e$ are designed to be orthogonal, capturing independent dimensions of fairness related to performance and cost, respectively.

Our proposed metrics also satisfy several desirable properties established in economic and computer science literature, including envy-freeness (FP.1), Pareto optimality (FP.2), fairness axiom (FP.3), and reduction (FP.4)~\cite{fairpro1,fairpro2,envyfree1,pareto,axiom,fair-eco}.
FU algorithms minimizing $M_p$ satisfy FP.1, FP.2, and FP.3; FU algorithms minimizing $M_e$ satisfy FP.3 and FP.4.
These properties ensure the robustness and theoretical grounding of our metrics. We provide formal definitions and proofs that FedShard satisfies these properties in Appendix F.

\section{Experimental Results}~\label{sec:evaluation}
In this section, we conduct comprehensive experiments on FedShard (FS) regarding the following evaluation questions. In \S\ref{sec:exp_sfl}, we answer \textbf{Q1}: How does the sharded framework affect federated training performance?
In \S\ref{sec:exp_efficiency}, we answer \textbf{Q2}: How fast and efficiency-fair is FedShard's FU mechanism? 
In \S\ref{sec:exp_performance}, we answer \textbf{Q3}: How effective and performance-fair is FedShard's FU mechanism?
For each subsection, we first explain the evaluation setting and metrics before presenting the results. 
We compare FS against state-of-the-art baselines spanning both calibration-based approximate methods—RapidRetrain (RR), FedEraser (FE), and FedRecovery (FR)—and exact methods—FATS (FT) and vanilla retraining via FedAvg (FA). More details like experimental errors, environment and settings are in Appendix G.1.

\subsection{Ablation Study: FedShard Performance}~\label{sec:exp_sfl}
\begin{figure}[ht]
    \centering
    \subfigure[]{
        \begin{minipage}[t]{0.48\linewidth}
            \centering
            \includegraphics[width=\textwidth]{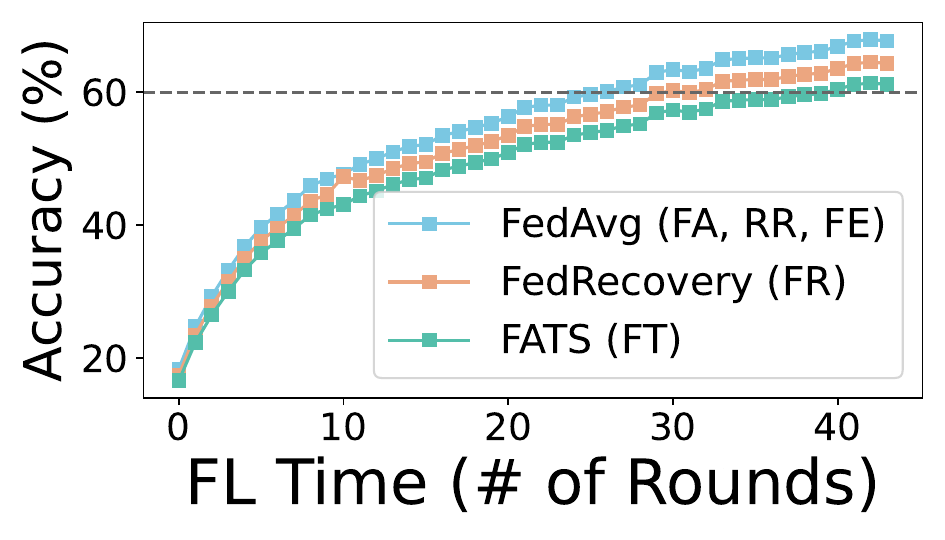}
        \end{minipage}%
    }
    \subfigure[]{
        \begin{minipage}[t]{0.48\linewidth}
            \centering
            \includegraphics[width=\textwidth]{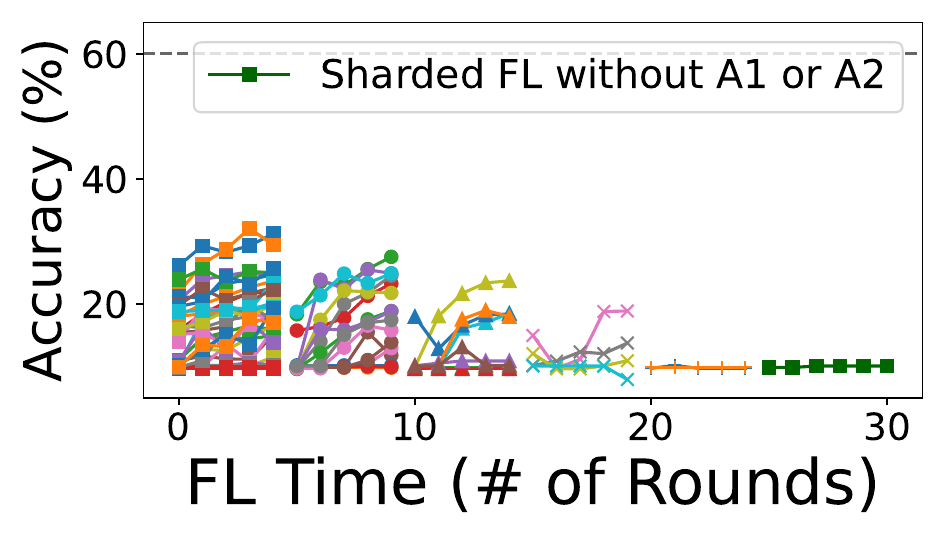}
        \end{minipage}%
    }
    \subfigure[]{
        \begin{minipage}[t]{0.48\linewidth}
            \centering
            \includegraphics[width=\textwidth]{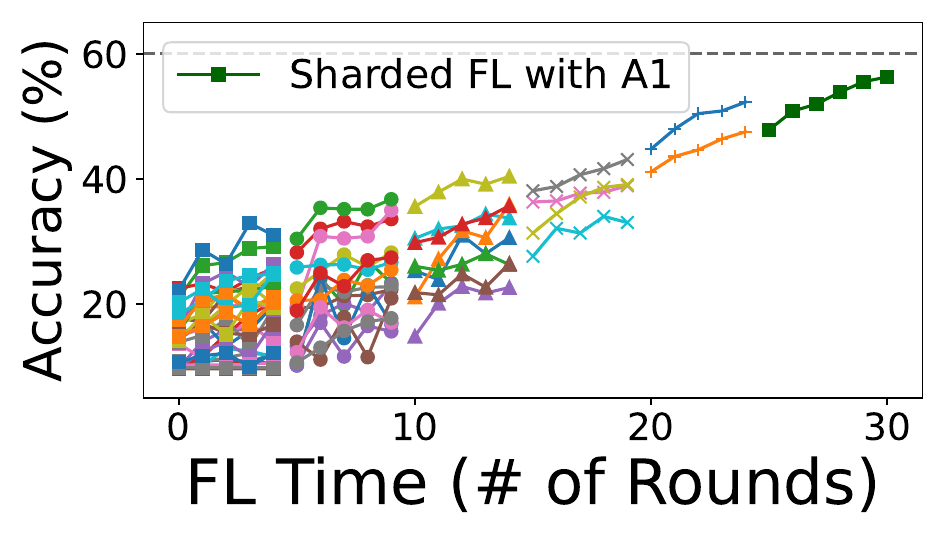}
        \end{minipage}%
    }
    \subfigure[]{
        \begin{minipage}[t]{0.48\linewidth}
            \centering
            \includegraphics[width=\textwidth]{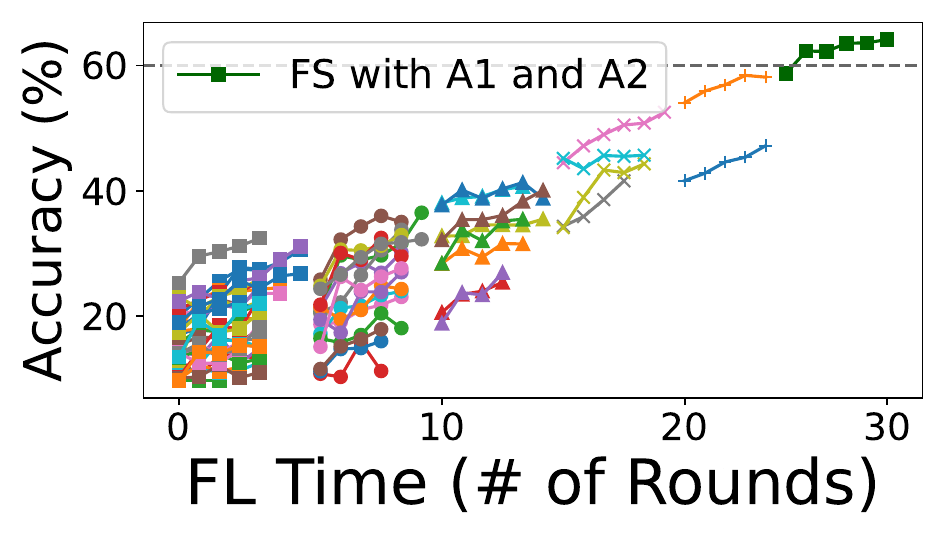}
        \end{minipage}%
    }
    \caption{Global model accuracy on CIFAR-10 (Non-IID $\rho=0.1$). The standard FedAvg (a) gives a baseline model performance on the federated training. The ablation study shows that naive sharding (b) is unstable. Our merging algorithm $\mathcal{A}_1$ (c) stabilizes training, and our round allocation algorithm $\mathcal{A}_2$ (d) further enables FedShard to surpass the accuracy of standard FedAvg.}\label{fig:acc}
\end{figure}

In this section, we evaluate FedShard's impact on the primary task of federated training. This experiment also serves as an ablation study, indicating the necessity of our proposed algorithms $\mathcal{A}_1$ (shard merging) and $\mathcal{A}_2$ (round allocation). We take CIFAR-10 with Non-IID degree $\rho=0.1$, the merging rate $R=2$ and the client number $K=32$ as an example, using a 2-layer CNN model as in FedAvg. 

Figure \ref{fig:acc} shows that our adaptive algorithms are crucial for achieving high model accuracy in a sharded environment. Firstly, as shown in Figure \ref{fig:acc}(a), the standard FedAvg achieves an accuracy of 62.92\% on CIFAR-10 using 30 epochs.
However, as shown in Figure \ref{fig:acc}(b), a naive sharded approach without our algorithms fails and suffers from severe model collapse issues. After merging of each stage, the accuracy of the new shards drops sharply, and the model struggles to converge in the new stage. This empirical result proves the necessity of our merging algorithm $\mathcal{A}_1$.

Second, as shown in Figure \ref{fig:acc}(c), our cluster-based merging algorithm $\mathcal{A}_1$ significantly mitigates this issue and stabilize the sharded training. By creating directionally balanced shards, $\mathcal{A}_1$ ensures stable convergence, though the accuracy (56.3\%) is still lower than standard FedAvg (62.92\%). With both $\mathcal{A}_1$ and $\mathcal{A}_2$, FedShard achieves an accuracy of 64.14\% as shown in Figure \ref{fig:acc}(d), even outperforming standard FedAvg. This further demonstrates the effects of $\mathcal{A}_2$.

This experiment fundamentally shows FedShard's practicality as a hierarchical federated training process, where $\mathcal{A}_1$ and $\mathcal{A}_2$ play an essential role in achieving even better convergence and model performance compared to traditional FL. Based on this, we further evaluate FedShard unlearning.

\subsection{Efficiency and Efficiency Fairness}~\label{sec:exp_efficiency}
Next, we evaluate FedShard's unlearning cost (time) and its E.F. ($M_e$ score). Due to space constraints, we show the experimental results on CIFAR-10 with Non-IID degree $\rho=0.1$ and 256 clients, while additional results for other datasets, Non-IID degrees, and client numbers are in Appendix G.2 and G.3.

Figure~\ref{fig:lc} illustrates the distribution of unlearning costs (\textit{i.e.}, unlearning time), revealing a clear efficiency and efficiency fairness dilemma among prior methods. Specifically, FA requires high unlearning costs. The state-of-the-art exact unlearning method (FT) exhibits extremely high variance of unlearning costs, signifying poor efficiency fairness. While approximate methods (FE, RR, and FR) are fast, they are challenged for performance unfairness as we later discuss in \S\ref{sec:exp_performance}. FedShard achieves low costs as calibration-based approximate methods (i.e., efficiency) while maintaining a fair cost distribution for all clients' unlearning requests (i.e., efficiency fairness).

\begin{figure}[ht]
    \centering
    \includegraphics[width=0.47\textwidth]{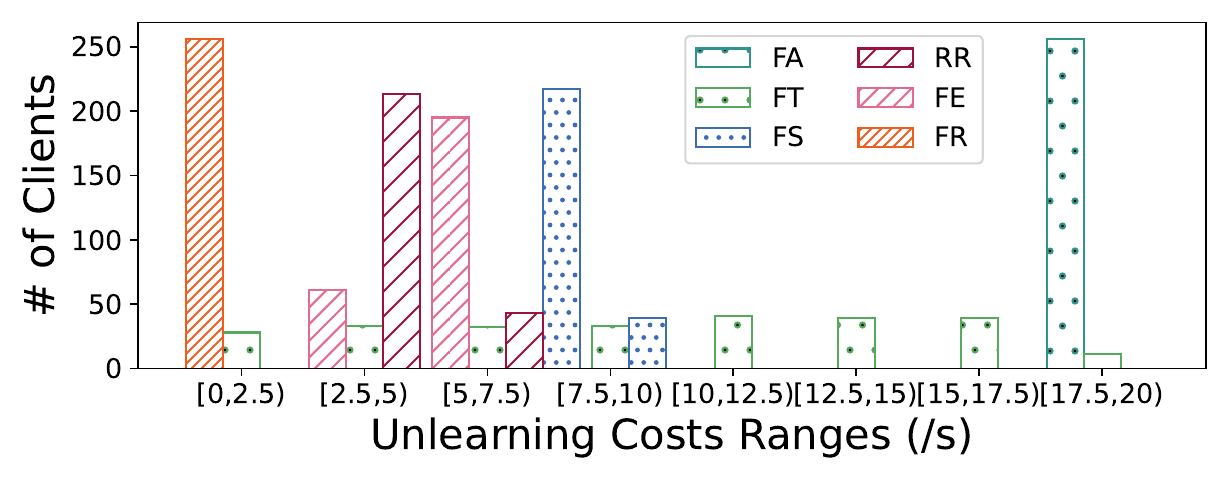}
    \caption{Unlearning cost distributions of different methods on CIFAR-10 with 256 clients, $R=2$ and $\rho=0.1$.}~\label{fig:lc}
\end{figure}

Table~\ref{tab:me} quantifies these results across multiple datasets though unlearning time $t$ and efficiency fairness metric $M_e$.
For the efficiency, FedShard is consistently one of the fastest methods, achieving up to a 6.2 times speedup over vanilla retraining (FA) and a 4.9 times speedup over the other exact method FT. For the efficiency fairness ($M_e$): FedShard consistently achieves the best or second-best $M_e$ score. In contrast, FT suffers from extremely high $M_e$ scores (e.g., 1081.73 on CIFAR-10), confirming its inherent efficiency unfairness. Approximate methods also exhibit worse E.F. than FedShard, especially under high Non-IID settings. We provide more experimental results in Appendix G.2, indicating that FedShard maintains its efficiency and E.F. robustly for various datasets, Non-IID degrees. Furthermore, with more clients, FedShard is more advanced in efficiency.

\begin{table}[ht]
    \centering
    \scriptsize

    \setlength{\tabcolsep}{1mm} 
    \begin{tabular}{|cc|c|c|c|c|c|c|}
        \hline
        && \multicolumn{1}{|c|}{FA} & \multicolumn{1}{c|}{FE} & \multicolumn{1}{c|}{RR} & \multicolumn{1}{c|}{FT}& \multicolumn{1}{c|}{FR} & \multicolumn{1}{c|}{\textbf{FS}}\\
        \hline

        \multirow{3}{*}{MNIST}  &$t$
    &\fbl{9244.6}         
    &3099.2       
    &2814.6      
    &6832.1    
    &\ftl{10.13}     
    &\ftl{2485.1}     \\
    & \tcl{$M_e$}  
    &\tcl{52.44}  
    &\sfttcl{35.18}  
    &\tcl{36.87}  
    &\fbtcl{727.66}  
    &\tcl{53.17}  
    &\sfttcl{9.51}      \\
    & \tdl{$M_p$} &\sfttdl{5.93}      
    &\tdl{9.57}      
    &\tdl{9.90}  
    &\tdl{6.54}  
   &\fbtdl{14.04}  
    &\sfttdl{6.08}      \\
    \hline

    \multirow{3}{*}{FMNIST}  &$t$   
    &\fbl{9984.4}        
    &2915.45       
    &\ftl{2214.2}      
    &6163.7     
    &\ftl{11.83}    
    &{2508.3}    \\
    &\tcl{$M_e$}   
   &\tcl{49.31}   
    &\tcl{32.64}   
     &\tcl{41.83}   
      &\fbtcl{447.59}   
     &\sfttcl{31.39}    
    &\sfttcl{11.62}    \\
    & \tdl{$M_p$}  
    &\sfttdl{5.85}        
    &\tdl{9.38}        
    &\tdl{9.82}  
    &\tdl{6.46}  
    &\fbtdl{14.23}  
    &\sfttdl{5.95}      \\

    \hline

    \multirow{3}{*}{CIFAR-10}  &$t$   
    &\fbl{13459.3}
    &4166.8
   &4644.7
    &8403.9
    &\ftl{17.34}
    &\ftl{3361.1}      \\
    &\tcl{$M_e$}   
    &\tcl{75.64}
    &\tcl{41.32}
    &\tcl{39.30}
    &\fbtcl{1081.73}
    &\sfttcl{35.92}
    &\sfttcl{10.9}     \\
    &\tdl{$M_p$}
    &\sfttdl{6.35}
    &\tdl{9.78}
    &\tdl{9.23}
    &\tdl{6.67}
    &\fbtdl{19.06}
    &\sfttdl{6.08}      \\

    \hline

    \multirow{3}{*}{CIFAR-100}  &$t$   
    &\fbl{29660.2}         
    &7576.1       
    &\ftl{6398.2}      
    &16822.5     
    &\ftl{18.01}    
    &{6859.2}      \\
    &\tcl{$M_e$}   
    &\tcl{76.98}   
    &\tcl{42.24}   
    &\tcl{41.58}   
    &\fbtcl{1103.06}    
    &\sfttcl{31.82}   
    &\sfttcl{11.23}\\
    &\tdl{$M_p$}  
   &\sfttdl{5.68}
    &\tdl{11.74}
    &\tdl{10.94}  
    &\tdl{7.13}  
    &\fbtdl{22.14}  
    &\sfttdl{5.94}      \\
    \hline
    \end{tabular}
    \caption{Average unlearning costs $t$ (/s), $M_e$ values, and $M_p$ values of different FU methods. This table shows the results with client number $K=512$ and non-iid level $\rho=0.1$. We bold the best two scores and underline the worst score. We provide the experimental error in Table 4 in Appendix G.2.}~\label{tab:me}%
\end{table}

    \subsection{Effectiveness and Performance Fairness}~\label{sec:exp_performance}
    Finally, we evaluate FedShard's effectiveness and performance fairness through membership inference attacks (MIA), a cascaded leaving simulation, and data poisoning attacks (DPA) specific to unfair FU.
    
    \begin{table}[ht]
        \centering
        \scriptsize
        \setlength{\tabcolsep}{1mm}
        \begin{tabular}{|c||c||c|c|c|c|c|c|}
            \hline
            \textbf{Method} & N.U & FA & FE & RR& FT&FR& \textbf{FS}\\
            \hline
            \textbf{MIA F1-Score}&{91.37} &34.18 &30.14 & {28.65}& 36.07&{25.07}&{35.04}\\
            \textbf{DPA Precision (\%)}&NaN&\sftl{0\%} &20.83\% &22.91\% & \sftl{0\%}& {28.50\%}&\sftl{0\%}\\
            \textbf{$M_p$ Score}&NaN& {6.23} & 9.81& 9.14 & 6.42& {21.55}&\sftl{5.95}\\
            \hline
        \end{tabular}
        \caption{MIA attacks and DPA attacks of different methods on CIFAR-10. N.U. refers to the non-unlearned model, serving as an MIA baseline.}~\label{tab:mia}
    \end{table}
    
    \textbf{Effectivenss evaluation via MIA attack experiment.} As shown in Table~\ref{tab:mia}, the FS effectively unlearns the data, reducing the MIA F1-score to a low level same as retraining FA. This is aligned with our theoretical analysis in Proposition \ref{prop:1}. Approximate methods (FE, RR, FR) often ``over-forget'', resulting in even lower F1-scores that hint at unfair performance degradation on remaining clients' data.
    
    \begin{figure}[ht]
        \centering
        \includegraphics[width=0.47\textwidth]{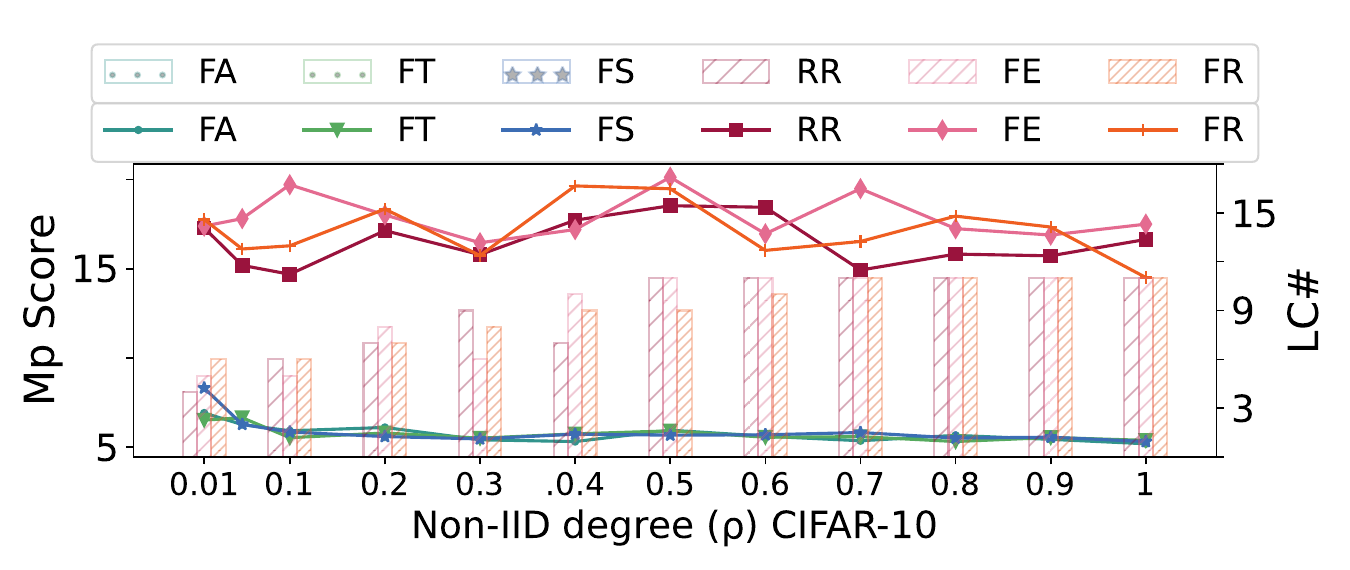}
        \caption{Comparison of $M_p$ and cascaded leaving. The lines (left y-axis) represent the scores of P.F. metric $M_p$ for different FU methods under different $\rho$ and the bars (right y-axis) show the number of cascaded leaving clients (LC\#).}~\label{fig:cl}
    \end{figure}
    \textbf{P.F. evaluation} \textbf{via} \textbf{cascaded leaving} \textbf{experiment}. 
    Cascaded leaving is a common issue in federated unlearning, where clients can leave the system due to unfair performance degradation caused by unlearning. The details and setting of the experiment are in Appendix G.5.
    
    Figure~\ref{fig:cl} demonstrates the P.F. of our FS through both the P.F. metric ($M_p$) and the number of cascaded leavers. 
    Specifically, calibration-based approximate unlearning methods including RR and FE have high $M_p$ scores (\textit{i.e.}, unfairer) and more cascaded leaving clients than retraining-based exact unlearning methods FS, FA, and FT. Their poor performance fairness causes cascaded leaving, and the number of cascaded leaving clients increases with non-IID degree decreasing. That is because with less non-IID data, the unfair performance degradation $\Delta \mathcal{Y}({D}_r)$ of approximate unlearning methods is more significant. Other retraining methods including FT and FA also achieve low $M_p$ scores and do not cause cascaded leaving as our proposed FS does. However, as discussed in \S~\ref{sec:exp_efficiency}, those methods are less efficiency-fair and less efficient compared with our FS.
    
    \textbf{P.F. evaluation via DPA Attack experiment.} 
    If an FU algorithm is not performance-fair (insufficient unlearning or over-unlearning), it can be easily attacked by data poisoning attacks (DPA)~\cite{dpa1,dpa2,dpa3,dpa4} as we explain in Appendix G.5. As shown in Table~\ref{tab:mia}, the approximate unlearning methods RR and FE fail to defend against DPA and exhibit high $M_p$ scores, consistent with the performance unfairness observed in the cascaded leaving experiment. In contrast, the retraining methods FS, FT, and FA successfully defend against DPA while achieving low $M_p$ scores, indicating that FS performs well in terms of P.F..

\subsection{Summary of Comparison Results}
\begin{figure}
    \centering
    \includegraphics[width=0.47\textwidth]{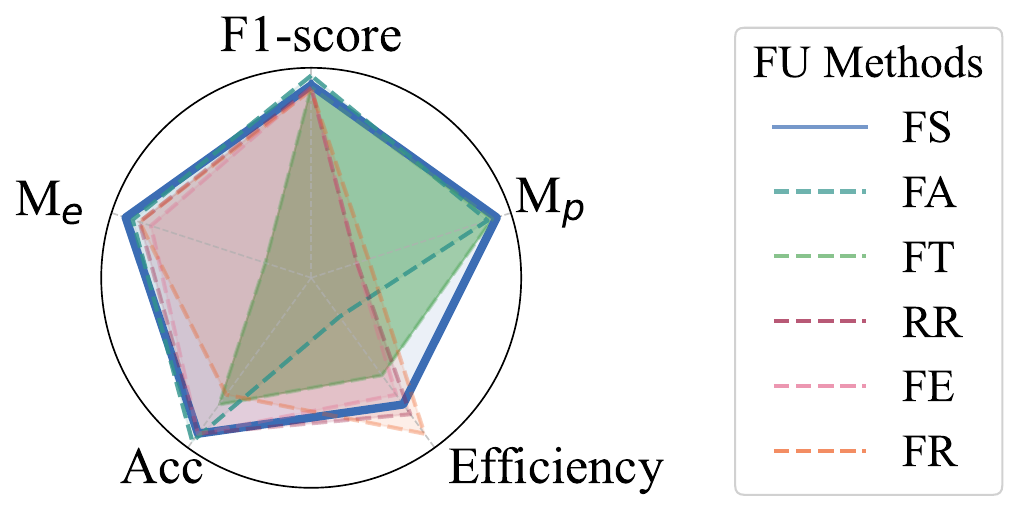}
    \caption{Holistic comparison of FU methods. Only FedShard (FS) excels across all five dimensions: Efficiency, efficiency fairness ($M_e$), performance fairness ($M_p$), model accuracy (Acc) and effectiveness (F1-score).}~\label{fig:fair}
\end{figure}

We summarize the results of the above three subsections in Figure~\ref{fig:fair}. Our experiments reveal a fundamental trade-off in prior work: approximate methods sacrifice P.F. for efficiency, while prior exact methods sacrifice E.F. or are prohibitively slow. FedShard is the only framework demonstrated to resolve this trade-off. As summarized in Figure~\ref{fig:fair}, FedShard uniquely achieves high efficiency, efficiency fairness, and performance fairness simultaneously.

\section{Conclusion}
In this paper, we study the performance fairness and efficiency fairness of federated unlearning. We propose two practical fairness metrics to measure the fairness degree of an unlearning method. More importantly, we introduce FedShard, a sharded federated learning and unlearning framework to achieve fair and efficient federated unlearning. Our theoretical analysis shows that FedShard achieves both performance fairness and efficiency fairness at least ${P}/{2}$ times faster unlearning than retraining from scratch, where $P$ is the stage number. Our evaluation results demonstrate that unfair federated unlearning methods cause cascaded leaving, poisoning attacks, and extreme unlearning cost distributions, which can be avoided by our fair FedShard unlearning.

\section{Acknowledgments}
This work is supported by National Natural Science Foundation of China under Grant 62502412. This research is also partially supported by National Key R\&D Program of China under Grant 2024YFE0200500 and National Natural Science Foundation of China under Grants 62202427 and 62572433.

\bibliography{aaai2026}

\setcounter{secnumdepth}{2}
\clearpage
\appendix
\section{Supplemental Related Work}~\label{seca:related}

\subsection{Fairness in Federated Unlearning}\label{seca:relatedfair}
While fairness is a well-established concern in machine learning and FL, existing research has focused on different axes of fairness than those we address.

Existing literature focused on data fairness and model fairness, which pay attention to whether specific groups of people suffer from racial discrimination~\cite{ful,fairFUL,dfairML2,dfairML3,fairMU} or whether data providers with different data qualities receive the same rewards~\cite{f2ul}. These notions of fairness are orthogonal to the concepts we introduce in this paper. Our work is the first to formalize and address fairness within the unlearning mechanism itself, focusing on the equitable distribution of performance impacts (P.F.) and computational burdens (E.F.) across the client population.

\subsection{Sharded Unlearning}\label{seca:relatedsharded}
The concept of sharding, or partitioning data to facilitate unlearning, has been explored previously in different contexts. SISA (Sharded, Isolated, Sliced, and Aggregated)~\cite{SISA} introduced a sharded architecture for centralized machine learning, where data is divided into mutually exclusive shards, and retraining is confined to only the shard containing the data to be forgotten. However, its reliance on a central server with direct data access makes it incompatible with the privacy-preserving constraints of FL.

More recently, ShardingEraser~\cite{ShardingE} applied sharding to FL and utilized sharding to accelerate an approximate unlearning method (FedEraser). Consequently, it inherits the fundamental limitations of approximate approaches and fails to guarantee performance fairness. In contrast, FedShard is the first framework to leverage a novel hierarchical sharding structure to achieve exact unlearning in the FL setting, while co-optimizing for both performance and efficiency fairness.

\subsection{Other Related Work}\label{seca:relatedhfl}
Hierarchical federated learning introduce multiple aggregation layers according to the physical network structure to reduce communication costs and improve scalability~\cite{hfl,hfl1}. FedShard, however, employs a logical hierarchical sharding structure to facilitate efficient and fair unlearning, which is orthogonal to the physical hierarchy in HFL. Moreover, original HFL's network/topology-based structure damages the E.F. required by FedShard, \textit{i.e.}, leaving clients on the critical path inevitably incurs high costs.

Clustered federated learning handle non-IID data by partitioning clients into clusters based on similarity and then training separate models for each cluster~\cite{cfl1}. However, FedShard's target is different in clustered federated learning scenarios, which lacks a unified global model for exact unlearning.

There are also machine unlearning literature on tree-based structures~\cite{DynFrs}. However, they focus on machine unlearning for random forest, where the tree-like structure serves as the model structure rather than FedShard, where a data partitioning strategy for federated unlearning.

\section{List of Symbols}~\label{seca:symbols}
We provide a list of symbols used in this paper for quick reference, including notations used in federated training and unlearning, fairness metrics, FedShard design and analysis.

\begin{table}[h]
    \centering
    \begin{tabular}{ll}

        \hline

        \textbf{Symbol}&\textbf{Description} \\

        \hline
        $c\in\mathcal{C}$&Federated clients, Set of all clients\\
        $\mathcal{C}_L, \mathcal{C}_R$&Set of leaving clients/ remaining clients\\
        $c_l,c_r$& certain leaving client/ remaining client \\
        $K$&Total number of clients ($K=|\mathcal{C}|$) \\
        ${D}_c\in\mathcal{D}$& Dataset of client $c$, Set of all datasets\\
        $\mathcal{Y}_{(D_c)}$&Model prediction accuracy of dataset $D_c$ \\
        $\mathcal{A}_{x}$&Algorithm \\
        $\theta, \Delta\theta$&Model parameters, $\theta$'s update \\
        $w_{c}, w_{s}^p$&Aggregation weight of client $c$/ shard $S_s^p$\\
        $\mathcal{L}_c(\theta, D_c)$&Loss function of client $c$ on dataset $D_c$ \\
        $\rho$& Dirichlet non-IID Level \\
        \hline
        $\alpha_{c}, \alpha_{s}^p$&Contribution factor of client $c$/ shard $S_s^p$ \\
        ${\rm Dis}(c_l,c_r)$&Distance between client $c_l$ and $c_r$ \\
        $Z_c, Z_{avg}$&Unlearning cost for client $c$/ for average\\
        $\sigma^2(\cdot)$&Variance of certain value \\
        \hline
        $R$&Merging rate ($R^{P-1}<K\le R^{P}$)\\
        $P$&Number of Stage\\
        $S_s^P\in\mathcal{S}^p$&Shard $s$ in stage $p$, Set of $S_s^P$ \\
        $N_p=|\mathcal{S}^p|$&Number of shards in stage $p$ \\
        $\mathcal{C}_s^p$&Set of clients in shard $S_s^p$ \\
        $T_s^p$&Training rounds of shard $S_s^p$ \\
        $\theta_s^p$&Model parameters of shard $S_s^p$ \\
        $\theta_s^{p,t=0}$&Initial model parameters of shard $S_s^p$ \\
        \hline
        $T_{max},T_{min}$&Max/Min training rounds of all shards \\
        $T_{train}, T_{un}$&Total training rounds for FL/FU \\
        $r_1$&The ratio of $T_{train}/T_{un}$\\
        $\hat{r_1}$&Consider $r_1$ with different $T_s^p$ \\
        $r_2$&Consider $r_1$ with multiple unlearning\\
        \hline
    \end{tabular}
    \caption{List of Symbols.}\label{tab:symbols}
\end{table}

\section{Contribution Factor}~\label{seca:contribution}
In this section, we discuss our choice of contribution factor $\alpha_c$ considering the trade-off between computational overhead and representativeness of the client contribution. We further discuss the alternatives from related works in FL personalization and adaptive aggregation, where measuring the data relevance of different clients are also required metrics.

In Equation (\ref{eq:dis}), we utilize the parameter similarity to measure the contribution factor $\alpha_c$ for every client, which is widely used in FL literature for adaptive aggregation~\cite{fedadp} and personalization~\cite{FedAMP}. With the cosine value of the model parameters, we capture adequate information about the data contribution of each client while having a small computational cost. 

The contribution factor can also be measured by the loss similarity.
The loss similarity utilizes the loss value of the model to measure the contribution factor of each client. For instance, we calculate the loss values on every client's local model. Lower loss value indicates higher contribution. This rough but fast method has already been used in personalization~\cite{FedFomo}. As a metric for the contribution factor, while this can be marginally faster to compute than vector similarity in then existing design, it is a scalar metric and thus less expressive. Two clients could reduce the loss by a similar amount but through conflicting updates; this critical nuance is captured by parameter similarity but lost when using only the loss value.

The Shapley value is an axiomatic method from cooperative game theory that provides a theoretically sound and provably fair attribution of contribution to each client~\cite{shapley}. This is also leveraged for personalization works to determine the data relevance~\cite{pFedSV}. The Shapley value is almost the theoretical ``gold standard'' for fair contribution attribution. However, its exact computation is NP-hard, and even state-of-the-art approximations are prohibitively expensive in large-scale FL systems. Its computational demands make it impractical for a dynamic and efficient framework like FedShard.

The information entropy utilizes information-theoretic measures to quantify a client's contribution, which has been proven to be practical in adaptive aggregation works~\cite{fedent}. This metric offers a different and valuable perspective on contribution, particularly by identifying clients that provide unique information. However, as a contribution factor, its computation is generally much more intensive than parameter similarity.

Considering the trade-off between low computational overhead and high representativeness of the client contribution, we choose the parameter similarity in the main paper. The other metrics can be used as alternatives for the contribution factor in FedShard, and we will explore the corresponding results in future work.

\section{Design Details and Algorithms}\label{seca:algo}
In this section, we provide more details about the design of FedShard using the same notations as in \S\ref{sec:fedshard}, which are also listed in Table \ref{tab:symbols}. We first elaborate the federated training process of FedShard in \S\ref{seca:algofed}, which serves as the foundation for the unlearning process. Then in \S\ref{seca:a1} and \S\ref{seca:a2}, we provides more details in the shard merging algorithm $\mathcal{A}_1$ and training round allocation algorithm $\mathcal{A}_2$, respectively.

\subsection{Federated Learning Algorithm of FedShard}\label{seca:algofed}

We first introduce the FL algorithm. For federated training, FedShard performs the following four phases in each stage $p$, where the first two phases are creating new shards and the last two phases are federated learning within the new shards:

\begin{algorithm}[h]
    \caption{FedShard Federated Learning Algorithm}
    \label{alg:a0}
    \textbf{Input}: Clients $\mathcal{C}=[c_1,c_2,\dots,c_K]$

    \textbf{Parameter}: Randomly initialized model $\theta^0$, Merging rate $R$, Empirical training rounds $T^*_0$ and $T^*_1$

    \textbf{Output}: Global model $\theta^*$, A cache for training process FLCache

    \begin{algorithmic}[1]
        
        \STATE $\alpha^0_i=0$, $w_i^0=1$, $\theta_i^0=\theta^0$, S$_i^0$ = $\mathcal{C}$[i] {\bfseries for} i=1, 2, $\dots$, K
        \STATE FLCache $\leftarrow$ S$_1^0$, S$_2^0$, $\dots$, S$_K^0$
        \STATE P=$\lceil \log_R K\rceil$, N$_0$=$K$
        \FOR{$p$ {\bfseries =} 1, 2, $\dots$,P}
            \STATE\textit{// Phase 1: Generating new shards}
            
            Shards = $\mathcal{A}_1$([S$_1^{p-1}$,\dots,S$_{N_{p-1}}^{p-1}$], [$\alpha_1^{p-1},\dots,\alpha_{N_{p-1}}^{p-1}$])
            
            \STATE N$_p$ = len(Shards) //\textit{Shards is a list [S$_1^p$, S$_2^p$, $\dots$, S$_{N_p}^p$]}

            \STATE\textit{// Phase 2: Initializing new shards}
            \STATE $\sigma^2$($\alpha_s$) = {\bfseries var}([$\alpha_i^{p-1}$ {\bfseries for} $i$ {\bfseries in} S$_s^p$]), $s$ {\bfseries =} 1,$\dots$,N$_p$
            \STATE $\sigma^2_{max}$, $\sigma^2_{min}$ = {\bfseries max}($\sigma^2$($\alpha_s$)), {\bfseries min}($\sigma^2$($\alpha_s$))
            \FOR{$s$ {\bfseries =} 1, 2, $\dots$, N$_p$}
                \STATE $\theta_s^{p,0}$ = {\bfseries Aggregate}([$\theta_i^{p-1}$], [$w_i^{p-1}$]), i {\bfseries in} S$_s^p$
                \STATE $w_s^p$ = {\bfseries sum}($w_i^{p-1}$) {\bfseries for} $i$ {\bfseries in} S$_s^p$
                \STATE $T_s^p$ = $\mathcal{A}_2$([$\alpha_i^{p-1}$ {\bfseries for} i {\bfseries in} S$_s^p$], $\sigma^2_{max}$, $\sigma^2_{min}$)
            \ENDFOR

            \STATE\textit{// Phase 3: Federated training in parallel}
            \FOR{each $S_s^p$ {\bfseries in} stage $p$, s{\bfseries =} 1, 2, $\dots$, N$_p$}
                \STATE$\theta_s^p$={\bfseries FL}($\theta_s^{p,0}$, [clients$\in S_s^p$], $T_s^p$)
            \ENDFOR

            \STATE\textit{// Phase 4: Update caching}
            \STATE $\theta^p$ = {\bfseries Aggregate}([$\theta^p_1$, $\theta^p_2$,\dots,$\theta^p_{N_p}$], [$w^p_1$, $w^p_2$,\dots,$w^p_{N_p}$])
            \STATE $\alpha_s^p$ = {\bfseries arccos}($\theta_s^p$,$\theta^p$) {\bfseries for} $s$ {\bfseries =} 1, 2, $\dots$, N$_p$
            \STATE FLCache $\leftarrow$ $S_s^p$, $\theta_s^p$, $\alpha_s^p$, $w_s^p$, $T_s^p$ for $s$ {\bfseries =} 1, 2, $\dots$, N$_p$

        \ENDFOR
    \end{algorithmic}
    \textbf{Return}: $\theta^*=\theta_1^P$, UpdateCache

    \textbf{Aggregate}(Thetas[],Weights[]):
    \begin{algorithmic}[1]
        \STATE $\theta$ = {\bfseries sum}($\theta_i \cdot w_i$) {\bfseries for} i {\bfseries =} 1, 2, $\dots$, N
        \STATE $\theta^*$ = $\theta$ / {\bfseries sum}($w_i$) {\bfseries for} i {\bfseries =} 1, 2, $\dots$, N
        \STATE \textbf{Return} $\theta^*$
    \end{algorithmic}
\end{algorithm}

\textit{(1) Generating new shards:} Firstly, the server merges previous shards $\{S_1^{p-1},S_2^{p-1},\dots,S_{N_{p-1}}^{p-1}\}$ from last stage to generate new shards $\{S_1^{p},S_2^{p},\dots,S_{N_{p}}^{p}\}$ with a shard merging algorithm $\mathcal{A}_{1}$:
\begin{equation}\label{eq:a1}
    \mathcal{A}_{1}:\mathcal{S}\rightarrow \{\mathcal{S}_1,\mathcal{S}_2,\dots,\mathcal{S}_{N}\},
\end{equation}
where $\cup_{\mathcal{S}_i}=\mathcal{S}$. The $\mathcal{A}_{1}$ outputs a partition on set $\mathcal{S}$, which indicates the shard merging strategy that all shards belonging to the same partition will merge as a new shard. For example, in Figure~\ref{fig:arch2}, previous shards $S_1^1,S_2^1,S_3^1$ are merged into new shard $S_1^2$ by $\mathcal{A}_{1}$.  For the new shards in the first stage, they are initialized by randomly picking up clients.

\textit{(2) Initializing new shards:} Then, the central server initializes all components for the new shards $S_s^p$, including the model parameters $\theta^{p,0}_s$, clients, and training rounds $T^{p}_s$. First, it sets the global model for the new shard $S_s^p$ by aggregating the global models from previous shards $S_i^{p-1}$:
\begin{equation}\label{eq:wa_sm}
    \theta^{p,0}_s \leftarrow \frac{1}{\sum_{i}   w_i^{p-1}} \sum_{i}   w_{i}^{p-1}\cdot\theta^{p-1}_{i},
\end{equation}
where $s_i^{p-1}$ refers to the shards that are merged into shard $S_s^p$. This is a classical aggregation process in FL like FedAvg, where the $w_i^{p-1}$ is the aggregation weight of shard $S_i^{p-1}$ and $\theta^{p-1}_{i}$ is the global model of shard $S_i^{p-1}$.
For example, in Figure~\ref{fig:arch2}, the global model of $S_2^2$ is initialized by aggregating the models of $S_4^1,S_5^1,S_6^1$. Then the union of clients from previous shards yields the clients of the new shards. In this example, $S_2^2$ contains $c_{7-14}$ and $c_{25}$, which come from the clients in $S_4^1$, $S_5^1$ and $S_6^1$. Besides, the central server sets the training rounds $T^p_s$ for each new shard by a training round assignment algorithm $\mathcal{A}_{2}$:
\begin{equation}\label{eq:a2}
    \mathcal{A}_{2}:{\{\alpha_c\}}\times{\{\mathcal{Y}\}}\times\mathbb{N}\rightarrow \mathbb{N}.
\end{equation}
This algorithm determines the training rounds $T^p_s\in\mathbb{N}$ for each new shard based to the contribution factors $\alpha_c$, model performance $\mathcal{Y}^{p-1}_s$, and previous training rounds $T^{p-1}_s$. 

\textit{(3) Federated Training:} After initializing, each new shard $S_s^p$ aims to decrease the average loss of its clients: \begin{equation}
    \min_{\theta_s^p} \frac{1}{|\mathcal{C}_s^p|}\sum_{c\in\mathcal{C}_s^p} \mathcal{L}_c(\theta_s^p, D_c),
\end{equation}
where $\mathcal{L}_c(\cdot)$ is the loss function for client $c$. To achieve this, each shard starts federated training among its clients $\mathcal{C}_s^p$ for $T_s^p$ rounds, with the initial global model $\theta^{p,0}_s$. In this work, we use FedAvg~\cite{fedavg} as the federated training algorithm, which is widely used in FL.

\textit{(4) Update Caching:} When a federated shard $S_s^p$ completes its training, it uploads the global model $\theta_s^p$ to the central server for cache. The central server further tests the model performance $\mathcal{Y}^{p-1}_s$ and updates the contribution factor $\alpha_c$ of each client.

\textbf{FLCache}: We cache the model parameters $\theta_s^p$ of all shards and all the hyperparameters during the training process, including the contribution factor $\alpha_s^p$, the aggregation weight $w_s^p$ and the shard structure $S_s^p$. The shard structure $S_s^p = [S_i^{p-1}]$ is a list of lists, which records the merging process. We can know which shards from the previous stage $p-1$ consist certain shard in stage $p$. 

The storage consumption of FLCache is tolerable, including the hyperparameters and model parameters. All the hyperparameters, such as training rounds $T_s^p$, aggregation weights $w_s^p$ and contribution factors $\alpha_s^p$ are scalars and only occupy very little space compared with model parameters $\theta_s^p$. It is reasonable to only consider FedShard's extra storage consumption on the model parameters. FedShard's space complexity is $\mathcal{O}(K)$, while the classical FL algorithm FedAvg also takes $\mathcal{O}(K)$ space for storage. Therefore, our memory costs are acceptable for the server where the storage and memory are usually considered as adequate. Related analysis is provided in \S\ref{seca:spacecom}.
\subsection{Shard Merging Algorithm $\mathcal{A}_1$}~\label{seca:a1}
The core idea of $\mathcal{A}_1$ is to first cluster the shards based on their updating directions and then merge the shards from different clusters to balance the updating directions as well as unlearning costs. Here we provide the pseudocode for the shard merging algorithm $\mathcal{A}_1$ in Algorithm \ref{alg:a1}, which takes the shards from the previous stage as input, merges them into new super-shards with more clients, and outputs the super-shards for the next stage.

\textbf{Direction-aware clustering}: Specifically, the server first simulates global aggregation to calculate the average updating direction. After that, FedShard naturally divides all shards into three clusters. The shards in the cluster with angles approximately equal to the average directions are more likely to converge. The shards in cluste with large positive or negative angles are less likely to converge. However, their average direction after aggregation can be converging. For example in Figure \ref{fig:merge}, $S_9^1$ has a large negative angle and $S_8^1$ has la arge positive angle. Both of them have convergence directions far from the average. However, after aggregating $S_7^1$, $S_8^1$ and $S_9^1$ together, the merged shard $S_2^2$ is initialized with a promising convergence direction.\footnote{Based on the angle, the shards can be further divided into 5 or more clusters. However, this is not necessary as three clusters are already working.}  

\textbf{Cluster-based merging}: We partition the shards into different groups, then the shards in the same group will be merged into a super-shard for the training of the next stage. The partition follows the principle that each partition has a similar number of shards from the three clusters. This principle ensures that the aggregated angle is close to the average, so that the merged super-model can converge more efficiently. For example, in Figure~\ref{fig:merge}, $\mathcal{A}_1$ generates the partition on $\{S^1_{1-9}\}$ as $\{\{S^1_{1-3}\},\{S^1_{4-6}\},\{S^1_{7-9}\}\}$. Therefore, in the next stage, new shards $S^2_2$ is initialized by merging $S^1_{7}$, $S^1_{8}$, and $S^1_{9}$. The newly aggregated model for shard $S^2_2$ begins with similar directions to average, making it easier to converge on a representative global model.

\begin{algorithm}[h]
    \caption{Shard Merging Algorithm $\mathcal{A}_1$}
    \label{alg:a1}
    \textbf{Input}: Shards [S$_1^{p-1}$,\dots,S$_{N_{p-1}}^{p-1}$], [$\alpha_1^{p-1},\dots,\alpha_{N_{p-1}}^{p-1}$]

    \textbf{Parameter}: Merging rate $R$

    \textbf{Output}: SuperShards [S$_1^{p}$,\dots,S$_{N_{p}}^{p}$] for next stage $p$

    \begin{algorithmic}[1]
        \STATE clus = {\bfseries KMeans}([S$_1^{p-1}$,\dots,S$_{N_{p-1}}^{p-1}$],[$\alpha_1^{p-1},...,\alpha_{N_{p-1}}^{p-1}$])

        //\textit{clus = [clus1, clus2, clus3], each is a list of S$_i^{p-1}$}
        \STATE  SuperShards[i]= $\emptyset$, {\bfseries for} i {\bfseries =} 1, 2, $\dots$, $\lceil\frac{N_{p-1}}{R}\rceil$

        \WHILE{clus2 {\bfseries not} empty}
            \STATE //\textit{Allocate shards with small $|\alpha|$}

            ShardTemp = {\bfseries min}(SuperShards, key=len) 
            \STATE allocateChildShard(ShardTemp, clus2) 
        \ENDWHILE

        \WHILE{clus1 and clus3 {\bfseries not} empty}
            \STATE ShardTemp = {\bfseries min}(SuperShards, key=len) 
            \IF {{\bfseries avgAlpha}(ShardTemp) $\ge 0$}
                \STATE //\textit{Allocate a shards with negative $\alpha$}
                \STATE {\bfseries allocateChildShard}(ShardTemp, clus1) 
            \ELSIF {{\bfseries avgAlpha}(ShardTemp) $<0$}
                \STATE //\textit{Allocate a shards with positive $\alpha$}
                \STATE {\bfseries allocateChildShard}(ShardTemp, clus3) 
            \ENDIF
        \ENDWHILE
    \end{algorithmic}
    \textbf{Return}: [S$_1^{p}$,\dots,S$_{N_{p}}^{p}$] = SuperShards

    \textbf{allocateChildShard}(ShardTemp, Cluster):

    \begin{algorithmic}[1]
        \FOR{ChildShard {\bfseries in} Cluster}
            \STATE $M_e$={\bfseries calculateMe}(ShardTemp, ChildShard)
        \ENDFOR
        \STATE ChildShard = {\bfseries min}(Cluster, key=$M_e$)
        \STATE ShardTemp.push(Cluster.pop(ChildShard))
    \end{algorithmic}
\end{algorithm}

\subsection{Training Round Allocation Algorithm $\mathcal{A}_2$}~\label{seca:a2}
\textbf{Empirical training rounds $T^*_0$ and $T^*_1$:} For the ideal case, we need to know the convergence rate of each stage to dynamically determine the training round range for each shard. However, in practice, no information about the convergence rate of a stage can be obtained until the training of the stage ends. Therefore, we take a preliminary experiment to acquire an empirical range of training rounds, which is practical with less overhead. 

In detail, we first train several stages, where each shard is trained until convergence. Then we record the minimum and maximum training rounds required for convergence as $T^*_0$ and $T^*_1$. Empirical observations indicate that the range set based on convergence rates from the first two stages approaches the optimal range, rendering them sufficiently feasible for the challenge.\footnote{Using example in Figure \ref{fig:acc}(d) as an instance of the empirical observations. In Figure \ref{fig:acc}(d), we observe that shards in the initial two stages take 4-7 rounds to converge. Then we set the range of training numbers for shards in the following stage as [4,7]. This range set is sufficient, as we avoid overfitting and too many training rounds for the last stages, which satisfies the challenge of why we design $\mathcal{A}_2$.} 

\textbf{Variance-based round allocation:} With the range determined, we adjust the training rounds for each shard in the subsequent stages. Existing literature~\cite{fedadp,fedent,asFL2} demonstrates that adaptive assignment of training resources (e.g., learning rates, aggregation weights, etc.) to clients who contribute more is crucial for accelerating convergence in federated learning and unlearning.
Therefore, for the remaining stages, we determine the training rounds $T_s^p$ for each shard $s$ in the same stage $p$ based on the following formula:
\begin{equation}
    \frac{T_s^p-T^*_0}{T^*_1-T^*_0}=\left[\frac{\sigma^2(\alpha_s)-\sigma^2(\alpha)_{min}}{\sigma^2(\alpha)_{max}-\sigma^2(\alpha)_{min}}\right]^{-1}, 
\end{equation}
where $\sigma^2(\alpha_s)$ (defined in Eq.\ref{eq:dis}) represents the variance of all clients' contributions in shard $s$. The $\sigma^2(\alpha)_{min}$ and $\sigma^2(\alpha)_{max}$ are the minimum and maximum contribution variance across all shards. The key idea of Equation (\ref{sec:adatra}) is to allocate more training rounds to shards with lower contribution variance. The clients in such shards have similar update directions, so allocating more training rounds helps enhance model accuracy.

Here we provide the pseudocode for the shard merging algorithm $\mathcal{A}_2$ in Algorithm \ref{alg:a2}, which takes the shards from the previous stage as input, merges them into new super-shards with more clients, and outputs the super-shards for the next stage.
\begin{algorithm}[h]
    \caption{Shard Merging Algorithm $\mathcal{A}_2$}
    \label{alg:a2}
    \textbf{Input}: Alphas=[$\alpha_1^{p-1}$,$\dots$]; $\sigma^2_{max}$ and $\sigma^2_{min}$ in stage $p$

    \textbf{Parameter}: Empirical training rounds $T^*_0$ and $T^*_1$

    \textbf{Output}: Training rounds $T_s^p$

    \begin{algorithmic}[1]
        \STATE $\sigma^2$ = {\bfseries var}([$\alpha_1^{p-1}$,$\dots$])
        \STATE $T_s^p$ = ($\sigma^2_{max}$-$\sigma^2_{min}$)/($\sigma^2$-$\sigma^2_{min}$) * ($T^*_1$-$T^*_0$) + $T^*_0$
    \end{algorithmic}
    \textbf{Return}: $T_s^p$
\end{algorithm}

\section{Details Step in Analysis}~\label{seca:details}
In \S\ref{sec:analysis}, we analyzed the effectiveness and efficiency of FedShard unlearning. Here we provide the details of the proofs for all propositions.

\subsection{Effectiveness of FedShard unlearning}\label{appsec:proof4s5}
In this section, we prove that FedShard unlearning, which only trains the one shard containing the leaving client every stage, is as effective as retraining from scratch. Proposition~\ref{prop:1} is well-established in the literature~\cite{SISA,FATS,exactfun,ShardingE}, stating that for any isolated subsystem in machine learning, if it is unaffected by departing clients, we can directly use cached updates without retraining. We define that $Y$ is affected by $X$ if and only if changes in $X$ lead to changes in $Y$.

\begin{proof}
    Let denote the leaving client as $c_l$.
    
    Base case: For the first stage $p=1$, $c_l$ is selected and only selected by one shard (named as $S_l^1$). For the other shards $S_i^1$ ($i\ne l$), each of them has random initial model $\theta_i^{p=1,0}$ (not affected by ${D}_l$) and train on ${D}_i$ (not containing data from $c_l$). Utilizing cached updates for those shards as retrained models is effective.

    Inductive Hypothesis: Assume the statement holds for stage $p-1$, \textit{i.e.}, only retraining the shard $S_l^{p-1}$ containing $c_l$ is effective.

    Inductive Step: For stage $p$, $S_l^{p-1}$ is merged and only merged into one shard $S_l^p$. Firstly, none of the other shards $S_i^{p}$ ($i\ne l$) contains data from $c_l$ as its training data. Secondly, the initial model $\theta_i^{p,0}$ of $S_i^{p}$ is not affected by $c_l$ as it is not merged from $S_l^{p-1}$. Therefore, using cached updates for $S_i^{p}$ ($i\ne l$) as retrained models is effective.

    Conclusion: By induction, we conclude that FedShard unlearning is as effective as retraining from scratch.
\end{proof}
\subsection{FedShard Unlearning Efficiency}~\label{appsec:effu}
To calculate $r_1$ in Proposition~\ref{lemma:r1}, we need to analyze the training rounds for both FedShard federated learning and federated unlearning. For federated learning, there are $K$ clients in every stage, regardless of the number of shards into which they are divided. Each client trains $T_0$ rounds. Therefore, the total training rounds for training are:
\begin{equation}
    T_{train}=P\cdot T_0\cdot K
\end{equation}
For federated unlearning, one shard of clients will be retrained in every stage. With the merging rate $R$, the number of clients of shards in stage $p$ is $R^p$. The total training rounds for federated unlearning are:
\begin{equation}~\label{eq:unlearning}
    \begin{aligned}
        T_{un}=&\sum_{i=1}^{P} T_0 R^i\\
        =&T_0 \frac{R(R^{P}-1)}{R-1}\\
    \end{aligned}
\end{equation}
Since there is only one shard in the last stage, all shards will be merged into that shard. We have
\begin{equation}
    R^{P}\approx K.
\end{equation}
Thus, we can calculate the $r_1$ as:
\begin{equation}~\label{eq:r1}
    \begin{aligned}
        r_1=&\frac{T_{train}}{T_{un}}\\
        &=\frac{P T_0 K}{T_0 \frac{R(R^{P}-1)}{R-1}}\\
        &= \frac{R-1}{R}\frac{K}{K-1}{P}.
    \end{aligned}
\end{equation}
The client number $K$ is usually much larger in an FL system. Therefore, we can approximate $K/(K-1)$ as $1$. The minimal value of the merging rate $R\in\mathbb{Z}^+$ is $2$. In that case, $r_1=P/2$. When $R$ is larger, the $r_1$ will approach $P$.

Considering different training rounds for the shard $s$ in stage $p$, denoted as $T_s^p$, we calculate a more accurate $\hat{r_1}$ in Proposition~\ref{thrm:r1p}. First, we need to point out that the $T_0$ in this situation refers to the average of $T_s^p$. Therefore, we have consistent conclusion with the case in Proposition~\ref{lemma:r1}. The training rounds for federated learning keep the same:
\begin{equation}
    \begin{aligned}
        T_{train}=&\sum_{i=1}^{P}\sum_{s=1}^{N_i} T_s^p * R^i\\
        =&\sum_{i=1}^{P}(R^i\cdot\sum_{s=1}^{N_i} T_s^p)\\
        =&P T_0 K,
    \end{aligned}
\end{equation}
where $R^i$ is the number of clients for shards in stage $p$.
Let consider the worst case. For the worst case, the shards containing leaving clients have the most training rounds $T_{max}$. Moreover, all the other shard happen to take $T_{min}$ training rounds. Therefore, we have the following equation for the worst case:
\begin{equation}\label{eq:betaprime}
    T_{min}=T_0-\beta T_0,\quad T_{max}=T_0+(N_p-1)\beta T_0,
\end{equation} 
where $N_p=R^{P-p}$ is the number of shards in stage $p$. We use $\beta$ to represent how much the minimal and maximal devirate from the average training rounds $T_0$. It is noted that $\beta=1-T_{min}/T_0 \in [0,1)$. A lower $\beta$ indicates a smaller variance of training rounds.
Considering that there is only one shard in the final stage, it must contain the leaving client and retrain for $T_0$ rounds. Therefore, the total training rounds for federated unlearning are:
\begin{equation}
    \begin{aligned}
        \hat{T}_{un}=&\sum_{i=1}^{P}T_l^i * R^i\\
        =&T_0R^P+\sum_{i=1}^{P-1}T_l^i * R^i\\
        \le&T_0K+\sum_{i=1}^{P-1}T_{max} * R^i,\\
    \end{aligned}
\end{equation}
where $S_l^i$ contains the leaving client in stage $i$. For the worst case, $T_{max}$ is in Equation~\ref{eq:betaprime}. Therefore, we have:
\begin{equation}
    \begin{aligned}
        \hat{T}_{un}\le&T_0K+\sum_{i=1}^{P-1}R^i*T_{max}\\
        =&T_0K+\sum_{i=1}^{P-1}R^iT_0*\left(1+R^{P-i}\beta-\beta\right)\\
        \le&T_0K+(1-\beta)\frac{K}{R-1}T_0+\sum_{i=1}^{P-1}R^iT_0*R^{P-i}\beta\\
        =&\frac{R-\beta}{R-1}T_0K+\beta(P-1)T_0K.\\
    \end{aligned}
\end{equation}
Based on the above equation, we can calculate the $\hat{r_1}$ as:
\begin{equation}
    \begin{aligned}
        \hat{r}_{1}=&\frac{T_{train}}{\hat{T}_{un}}\\
        =&\frac{P T_0 K}{\frac{R-\beta}{R-1}T_0K+\beta(P-1)T_0K}\\
        =&\frac{P}{\frac{R-\beta}{R-1}+\beta(P-1)}\\
        \approx&\frac{T_0}{T_{max}}\cdot r_1.
    \end{aligned}
\end{equation}
If $T_{max}=T_{min}$, \textit{i.e.}, $\beta=0$, we have $\hat{r}_{1}=P/(R-1)$. If $\beta$ is small, \textit{i.e.}, the bound of $T_s^p$ is small, we have $\hat{r}_{1}= PR/(R-1)$, which is aligned with $r_1$. For the worst case $\beta\rightarrow 1$, the $\hat{r}_{1}\rightarrow 1$, meaning that unlearning is as costly as retraining. 
This indicates that only if we maintain a narrow bound of $T_s^p$, the efficiency of FedShard unlearning will be guaranteed. 

For the best case, where the retrained shards perform the minimal training rounds, we have:
\begin{equation}
    \begin{aligned}
        \hat{T}_{un}=&\sum_{i=1}^{P}T_s^i * R^i\\
        =&\sum_{i=1}^{P}T_{min} * R^i\\
        =&T_{min}\cdot \frac{R(R^{P}-1)}{R-1}.\\
    \end{aligned}
\end{equation}
In this situation, we have:
\begin{equation}
    \begin{aligned}
        \hat{r}_{1}=&\frac{T_{train}}{\hat{T}_{un}}\\
        \le&\frac{P\cdot T_0\cdot K}{T_{min}\cdot \frac{R(R^{P}-1)}{R-1}}\\
        =&\frac{T_0}{T_{min}}\cdot r_1.
    \end{aligned}
\end{equation}
This indicates an even better FedShard unlearning efficiency. Therefore, for not only prompting the unlearning efficiency but also guaranteeing the efficiency fairness, it is theoretically necessary to keep the training rounds of all shards in a narrow range.
\subsection{Multiple Unlearning Efficiency}~\label{appsec:effm}
For the best case where all $m$ leaving clients are distributed in the same shard in the first stage, there will only be one shard affected for all following stages. Unlearning them together is $r_2^+$ times faster than unlearning one by one, where:
\begin{equation}
    r_2^+=m,
\end{equation}
which is evident. For the worst case where all $m$ leaving clients are distributed in different shards for the first stage, the total unlearning rounds are $T_{mun}=T_1+T_2$. $T_1$ refers to the earlier stages, where the number of shards is more than leaving clients, so retraining can be bypassed partially:
\begin{equation}
    T_{1}=m\cdot\sum_{i=1}^{s'} T_0\cdot R^{i}   =m\frac{R^{s'+1}-1}{R-1} T_0 \\
             \leq m{R^{s'}}T_0
\end{equation}
Considering that $m{R^{s'}}$ is less than $K$, 
\begin{equation}T_1\leq KT_0.\end{equation}
\begin{equation}T_2=\sum_{i=s'+1}^{P} T_0\cdot K=(P-s'-1)KT_0\end{equation}
refers to the latter stages where all shards contain leaving clients. The total number of unlearning rounds is:
\begin{equation}
    T_{mun}=T_1+T_2\leq =(P-s')KT_0.
\end{equation}
The $s'$ is related to the value of $m$. Specifically, given $m$ clients leaving simultaneously, for the worst case $$s'=P-x,$$ where in this stage, the number of shards $R^x$ is equal to $m$. Therefore, we have $R^x = m$, \textit{i.e.}, $x= \lg_R m$ and $s'=P-\lg_R m$ for the worst case. As a result, we have 
\begin{equation}~\label{eq:mun_tc}
    \begin{aligned}
        T_{mun}&=T_1+T_2\\
        &\leq (P-\lg_R{m})KT_0\\
        &=(\lg_R{K}-\lg_R{m})KT_0\\
        &=(\lg_R{K/m})KT_0
    \end{aligned}
\end{equation}

That is to say, multiple unlearning is even $r_2$ times faster than rapid unlearning one by one, where:
\begin{equation}\label{eq:r2}
    r_2=\frac{m\cdot T_{un}}{T_{mun}}\ge\frac{R}{R-1}\frac{K-1}{K}\frac{m}{(P-s')}.
\end{equation} 
Therefore, we have:
$$
    r_2^-=\frac{R}{R-1}\frac{K-1}{K}\frac{m}{(P-s')}=\frac{R}{R-1}\frac{K-1}{K}\frac{m}{\lg_R{K/m}}\ge 1.
$$
\subsection{Time Complexity Analysis}~\label{appsec:timecom}
Firstly, we prove that the time complexity of FedShard unlearning is $\mathcal{O}(K)$ for one client. According to Eq.\ref{eq:unlearning}, the total training rounds for federated unlearning is:
$$
    T_{un}=\frac{T_0R(R^{P}-1)}{R-1}.
$$
Since the time for training one round can be considered as a constant, the time complexity of FedShard unlearning is $$\mathcal{O}(T_{un})=\mathcal{O}(\frac{T_0R(R^{P}-1)}{R-1})=\mathcal{O}(R^{P}).$$
Given that the merging rate $R$ is a constant and $P=\lg_R K$, we have:
$$\mathcal{O}(T_{un})=\mathcal{O}(R^{P}/R)=\mathcal{O}(K).$$
Then we prove that the time complexity of FedShard unlearning is $\mathcal{O}(K\lg m)$ for $m$ clients.
According to Eq.\ref{eq:mun_tc}, the total retraining rounds for FedShard is $T_{mun}=(P-\lg_R{m})KT_0$. Then we have:
$$\mathcal{O}(T_{mun})=\mathcal{O}((P-\lg_R{m})KT_0)=\mathcal{O}(K\lg m).$$

\subsection{Space Complexity Analysis}~\label{seca:spacecom}
In this section, we analyze the space complexity of FedShard unlearning. First of all, the space complexity of FedAvg, which is the basic algorithm of FedShard, is $\mathcal{O}(K)$, as it needs to store the model parameters for all $K$ clients during the aggregation process. It is noted that FedAvg will release the model parameters after the aggregation, which means that the space complexity is not related to the training round $T$.

In FedShard, we also need to store the model parameters for all $K$ clients during the aggregation process, which is $\mathcal{O}(K)$. FedShard also does not hold them permanently, so the space complexity is not related to the training round $T_s^p$ as well. However, in order to support the unlearning process, FedShard retains the final model parameters of every stage. In each stage $p$, the number of shards is $$N_p = \lceil\frac{K}{R^p}\rceil\le \frac{K}{R^p}+1.$$ The space consumption by model parameters is the same for shard or client, beacuse they are training the same model. To hold all shards in $P$ stages, we need additional storage:
$$\sum_{p=1}^{P} \frac{K}{R^p}+1\le K+P.$$
Considering that $P=\lg_R K$, the space complexity of FedShard unlearning is:
$$\mathcal{O}(K)+\mathcal{O}(K+\lg_R K)=\mathcal{O}(K),$$
where the first $\mathcal{O}(K)$ is the space complexity of clients and the second $\mathcal{O}(K+\lg_R K)$ is the additional space complexity caused by FedShard unlearning.
In conclusion, the space complexity of FedShard unlearning is $\mathcal{O}(K)$, which is the same as FedAvg.

\section{Properties of Fairness Metrics}\label{seca:properties}
In \S~\ref{sec:metrP}, we introduce that our proposed metrics $M_p$ and $M_e$ can measure the fairness of unlearning algorithms with many practical properties. Here, we provide the definitions of the properties and give details of the proofs for all propositions.

Our analysis starts with the properties of $M_p$, including envy-freeness, Pareto optimality, axiom of continuity, partition, homogeneity and starvation. Then we prove the properties of $M_e$, including reduction to equality, axiom of continuity, partition, saturation and starvation.

\subsection{Properties of $M_p$}
\begin{proposition}\label{prop:envy-freeness}
    \textbf{Envy-freeness}: Minimizing $M_p$ ensures that no client prefers another's leaving decision.
    \begin{proof}
        Let $\{a_c\}_{c=1}^K$ be the choices of federated clients after learning and $\{a_c^\prime\}_{c=1}^K$ be the choices after unlearning:
        $$a_c=\left\{\begin{aligned}
            -1, \quad&\text{if } U_c(-1) > U_c(1)\text{ and client $c$ leaves;}\\
            1, \quad&\text{if } U_c(1)\ge U_c(-1)\text{ and client $c$ stays.}\\
        \end{aligned}\right.
        $$
        Here $U_c(a_c)$ is the utility function of client $c$ with choice $a_c$, which is related to constant factors such as training costs and variable factors. 
        $$\left\{\begin{aligned}
                U_c(-1)\propto {\rm Dis}(c,c_r);\\
                U_c(1)\propto \Delta \mathcal{Y}({D}_c).
        \end{aligned}\right.$$
        The utility function of leaving is positively correlated with the contribution distance which indicates its data privacy concerns. The utility function of staying is related to model accuracy degradation, which influences their benifits from federated learning.

        For the leaving clients, envyness indicates that 
        $$\left\{\begin{aligned}
            U_c(-1) > U_c(1),\\
            U_c^\prime(-1)\le U_c(1).
        \end{aligned}\right.$$
        Consequently, we have $$U_c(-1)-U_c^\prime(-1) > U_c(1)-U_c(1),$$ meaning a small ${\rm Dis}(c,c_r)$ and a large $\Delta \mathcal{Y}({D}_c)$ during unlearning operation, which is a contradiction to minimizing $M_p$ scores.

        Similarly for the staying clients. Envyness here indicates that the clients preferred to stay after learning turn to prefer to leave after unlearning, which is consistent with the concept of \textit{cascaded leaving}. 
        This means a large ${\rm Dis}(c,c_r)$ and a small $\Delta \mathcal{Y}({D}_c)$ during unlearning operation, which is also a contradiction to minimizing $M_p$ scores.

        To conclude, minimizing $M_p$ avoids envyness during unlearning operations and ensures that no client prefers another's leaving decision.
    \end{proof}
\end{proposition}

\begin{proposition}
    \textbf{Pareto optimality}: The performance of unlearned model with minimal $M_p$ scores is Pareto optimal for all federated clients.
    \begin{proof}
        First, let prove that for any other performance degradation $\Delta\mathcal{Y}$ made by unlearning, if it is a Pareto improvement for one client $P.I(\Delta\mathcal{Y})$, then the minimal $M_p$ is surely not achieved.

        Let denote the original solution of performance degradation as $\Delta\mathcal{Y}$ and denote a Pareto improvement as $\Delta\mathcal{Y}^\prime$. Without generality, let
        \begin{equation}
            \begin{aligned}
            \min_{c_r\in\mathcal{C}_R}{\rm Dis}(c_1,c_r) &< \min_{c_r\in\mathcal{C}_R}{\rm Dis}(c_2,c_r)\\ &<\dots \\ &<\min_{c_r\in\mathcal{C}_R}{\rm Dis}(c_K,c_r) \\
            \end{aligned}
        \end{equation}
        According to Lemma~\ref{lemma:mpequiv2}, the minimal $M_p$ is achieved when the performance degradation $\Delta\mathcal{Y}$ caused by unlearning operations satisfies
        $$
        \Delta\mathcal{Y}({D}_{c_1}) < \Delta\mathcal{Y}({D}_{c_2}) < \dots < \Delta\mathcal{Y}({D}_{c_K}).
        $$
        The Pareto improvement requires a better off for one client without making another client worse off, indicating that 
        \begin{equation}
        \begin{aligned}
        \exists c_i, c_j, &s.t. \Delta\mathcal{Y}^\prime({D}_{c_i}) < \Delta\mathcal{Y}({D}_{c_i})\\ 
        & \text{ and } \Delta\mathcal{Y}^\prime({D}_{c_j}) \le \Delta\mathcal{Y}({D}_{c_j}).\\
        \end{aligned}
        \end{equation}
        That will definitely not lead to a minimal $M_p$ score. 

        As a result, we prove that
        $$
            \exists \Delta\mathcal{Y}^\prime, P.I.(\Delta\mathcal{Y}^\prime) \rightarrow \neg \min_{\mathcal{A}_{\text{fu}}\in \{\mathcal{A}\}}M_p,
        $$
        which is the equivalent to its contrapositive form
        $$
            \min_{\mathcal{A}_{\text{fu}}\in \{\mathcal{A}\}}M_p \rightarrow  \neg \exists \Delta\mathcal{Y}^\prime,  P.I.(\Delta\mathcal{Y}).
        $$
        According to the definition, if there does not exist a Parato improvement, the Pareto optimal is reached. Therefore, unlearning algorithms with minimal $M_p$ scores are Pareto optimal for all federated clients.
    \end{proof}
\end{proposition}

\begin{lemma}~\label{lemma:mpequiv2}
    Finding an unlearning algorithm minimizing $M_p$ is equivalent to finding an unlearning algorithm so that the performance degradation caused by unlearning is consistent with client distribution distance regarding order.
    \begin{equation}
        \begin{aligned}
            &\min_{\mathcal{A}_{\text{fu}}\in \{\mathcal{A}\}}M_p\\
            =&\min_{\mathcal{A}_{\text{fu}}\in \{\mathcal{A}\}}\frac{1}{|\mathcal{C}|}\sum_{c\in\mathcal{C}} f_{\oplus}(\Delta \mathcal{Y}({D}_c),\min_{c_r\in\mathcal{C}_R}{\rm Dis}(c,c_r))\\
            \equiv& \min_{\mathcal{A}_{\text{fu}}\in \{\mathcal{A}\}}\sum_{c\in\mathcal{C}} |\Delta \tilde{\mathcal{Y}}({D}_c)-\min_{c_r\in\mathcal{C}_R}\tilde{{\rm Dis}}(c,c_r)|\\
            \equiv& \min_{\mathcal{A}_{\text{fu}}\in \{\mathcal{A}\}}\sum_{c\in\mathcal{C}} |{\rm Rk}(\Delta \mathcal{Y}({D}_c))-{\rm Rk}(\min_{c_r\in\mathcal{C}_R}{\rm {\rm Dis}}(c,c_r))|,
        \end{aligned}
    \end{equation}
    where ${\rm Rk}(x)$ is order for $x$ in the set $\{x\}$. The function $\tilde{x}$ here normalized $x$ into $(0,1]$.

    \begin{proof}
        For convenience of presentation, we use $a_i$ to replace $\Delta$$\mathcal{Y}$(${D}_{c_i}$) and use $b_i$ to replace $\min_{c_r\in\mathcal{C}_R}{\rm Dis}(c_i,c_r)$, where $i$=1, 2, $\dots$, K. Without generality, we assume that $b_1$ $\le$ $b_2$ $\le$ $\dots$ $\le$ $b_K$.

        First, the proof of the equivalence between the first and second lines of $M_p$ is straightforward. The function $f_{\oplus}(x,y)$ is a function that first decreases and then increases with a minimum point at $x=y$. When the minimals of all $f_{\oplus}(x,y)$ is achieved simultaneously, the minimal $M_p$ is also achieved.

        Then we prove $\min_{\mathcal{A}_{\text{fu}}\in \{\mathcal{A}\}}M_p$ equivalent to the third line. Given $i<j$ s.t. $b_i<b_j$ and $a_i<a_j$. If we change the value of $a_i$ and $a_j$, the change of $M_p$ will be $\Delta\cdot 1/|\mathcal{C}|$, where $\Delta$ is
        $$
        \begin{aligned}
            \Delta=&\left(f_{\oplus}(a_i,b_i) + f_{\oplus}(a_j,b_j)\right)-\left(f_{\oplus}(a_i,b_j) + f_{\oplus}(a_j,b_i)\right)\\
            =&\frac{(a_i+b_i)^2}{a_ib_i}+\frac{(a_j+b_j)^2}{a_jb_j}-\frac{(a_i+b_j)^2}{a_ib_j}-\frac{(a_j+b_i)^2}{a_jb_i}\\
            =&\frac{(a_i-a_j)(b_i-b_j)(a_ib_i-a_jb_j)}{a_ia_jb_ib_j} \\
        \end{aligned}
        $$
        Since $a_i<a_j$, $\Delta$ is positive, indicating a higher $M_p$ score. According to the principles of Rearrangement Inequality, the minimal $M_p$ is achieved when $a_i$ and $a_j$ are in the same order as $b_i$ and $b_j$ for any $i$ and $j$.
    \end{proof}
\end{lemma}

\begin{proposition}
    \textbf{Axiom of Continuity}: Fairness metrics $M_p$ is continuous with respect to the performance degradation and client distribution distance.
    \begin{proof}
        The continuity of $M_p$ is straightforward. According to the definition of $M_p$
        $$
            \begin{aligned}
                M_p \triangleq & \frac{1}{|\mathcal{C}|}\sum_{c\in\mathcal{C}} f_{\oplus}(\Delta \mathcal{Y}({D}_c),\min_{c_r\in\mathcal{C}_R}{\rm Dis}(c,c_r)),\\  &\text{where }f_{\oplus}(x,y)=(\tilde{x}+\tilde{y})\cdot(\frac{1}{\tilde{x}}+\frac{1}{\tilde{y}}),
            \end{aligned}
        $$
        the continuity of $M_p$ is guaranteed by the continuity of $f_{\oplus}(x,y)$, $\Delta \mathcal{Y}({D}_c)$, and ${\rm Dis}(c,c_r)$.
    \end{proof}
\end{proposition}

\begin{proposition}
    \textbf{Axiom of Homogeneity}: Fairness metrics $M_p$ is Fairness measure f(x) is a homogeneous function of degree 0, being independent of the unit of measurement
    or magnitude of the resource allocation.
    \begin{proof}
        The function $f_{\oplus}(x,y)=(\tilde{x}+\tilde{y})\cdot(\frac{1}{\tilde{x}}+\frac{1}{\tilde{y}})$ is a homogeneous function. Then we only need to prove that $\Delta \mathcal{Y}({D}_c)$ and ${\rm Dis}(c,c_r)$ are independent of the unit of measurement. That is obvious due to the normalization function:
        $$
            \tilde{x}=\epsilon_0+(x-x_{min})/(x_{max}-x_{min}+\epsilon_1),
        $$
        which normalizes $x$ into $(0,1]$ and makes $f_{\oplus}(x,y)$ independent of the unit of measurement.
    \end{proof}
\end{proposition}

\begin{proposition}
    \textbf{Against the Axiom of Saturation}: It is noted that the fairness metrics $M_p$ is not saturated, \textit{i.e.}, the value of $M_p$ varies with the number of clients in the system. We compare the performance fairness between federated unlearning algorithms by $M_p$ with the same federated learning settings including the same number of clients and the same dataset.
\end{proposition}

\begin{proposition}
    \textbf{Axiom of Partition}: Consider an arbitrary partition of all the clients into two groups. The $M_p$ score for total FL system is the mean of $M_p$ scores for two groups, for all possible partitions.
    \begin{proof}
        Let $\mathcal{C}_1$ and $\mathcal{C}_2$ be the two groups of clients satisfying $$\mathcal{C}_1\cup\mathcal{C}_2=\mathcal{C} \text{ and } \mathcal{C}_1\cap\mathcal{C}_2=\emptyset$$. The $M_p$ score for the total FL system is
        $$\begin{aligned}
            M_p&=\frac{1}{|\mathcal{C}|}\sum_{c\in\mathcal{C}} f_{\oplus}(\Delta \mathcal{Y}({D}_c),\min_{c_r\in\mathcal{C}_R}{\rm Dis}(c,c_r))\\
            &=\frac{1}{|\mathcal{C}_1|+|\mathcal{C}_2|}\left[\sum_{c\in\mathcal{C}_1} f_{\oplus}(\Delta \mathcal{Y}({D}_c),\min_{c_r\in\mathcal{C}_R}{\rm Dis}(c,c_r))\right.\\ & \left.+\sum_{c\in\mathcal{C}_2} f_{\oplus}(\Delta \mathcal{Y}({D}_c),\min_{c_r\in\mathcal{C}_R}{\rm Dis}(c,c_r))\right]\\
            &=\frac{\mathcal{C}_1|}{|\mathcal{C}_1|+|\mathcal{C}_2|}M_{p_1}+\frac{\mathcal{C}_2|}{|\mathcal{C}_1|+|\mathcal{C}_2|}M_{p_2}.\\
        \end{aligned}$$
        Therefore, for all possible partitions, the $M_p$ score for total FL system is the weighted average of $M_p$ scores for two groups, with the numbers of the clients as the weights.
    \end{proof}
\end{proposition}

\begin{proposition}
    \textbf{Axiom of Starvation}: $M_p$ is aligned with the result that starvation is no more fair than equal distribution.
    \begin{proof}
        For equal distribution in the performance degradation after federated unlearning, an FU algorithm results in equal performance degradation $\Delta\mathcal{Y}$({D}$_j$)$\neq$0 for all leaving clients $\mathcal{C}_0$ and equal performance degradation $\Delta \mathcal{Y}({D}_i)=0$ for all remaining clients $\mathcal{C}_1$. The $M_p$ score for an equal distribution is
        \begin{equation}\begin{aligned}
            M_p
            &=\frac{1}{|\mathcal{C}|}\sum_{c\in\mathcal{C}} f_{\oplus}(\Delta \mathcal{Y}({D}_c),\operatorname{MIN})\\
            &=\frac{|\mathcal{C}_0|}{|\mathcal{C}|}\left[2+\frac{1}{|\mathcal{C}_0|\epsilon_0}\sum_{c\in\mathcal{C}_0}\operatorname{MIN}\right]   \\
            &+\frac{|\mathcal{C}_1|}{|\mathcal{C}|}\left[2+\frac{1}{|\mathcal{C}_1|}\sum_{c\in\mathcal{C}_1}(\operatorname{MIN}+    \frac{1}{\operatorname{MIN}})\right],\\
        \end{aligned}
        \end{equation}
        where $\operatorname{MIN}$ represents $\min_{c_r\in\mathcal{C}_R}{\rm Dis}(c,c_r)$ for the convenience of notation.
        For starvation, we assume starvation without consideration on client contribution, which means that all performance degradation are the same $\Delta \mathcal{Y}({D}_i)=\Delta \mathcal{Y}({D}_j)$. Then the $M_p$ score for an equal distribution is
        $$\begin{aligned}
        M_p&=\frac{1}{|\mathcal{C}|}\sum_{c\in\mathcal{C}} f_{\oplus}(\Delta \mathcal{Y}({D}_c),\operatorname{MIN})\\
        &=\frac{1}{|\mathcal{C}|}\sum_{c\in\mathcal{C}} (\epsilon_0+\operatorname{MIN})(\frac{1}{\epsilon_0}+\frac{1}{\operatorname{MIN}})\\
        &=2+\frac{1}{|\mathcal{C}|\epsilon_0}\sum_{c\in\mathcal{C}} \operatorname{MIN}\\
        \end{aligned}$$

        Since the $\epsilon_0$ is a quite small value approaching 0, the $M_p$ score for starvation is larger than the $M_p$ score for equal distribution. Therefore, $M_p$ is aligned with the result that starvation is no more fair than equal distribution.
    \end{proof}
\end{proposition}

\subsection{Properties of $M_e$}
\begin{proposition}
    \textbf{Reduction to Equality}: Federated unlearning algorithm minimizing $M_e$ can reduce to equality if the clients have the same data contribution. 
    \begin{proof}
        We start the proof with the definition of $M_e$:
        $$M_e\triangleq \frac{1}{|\mathcal{C}|}\sum_{c\in\mathcal{C}} \frac{(Z_{avg}-Z_c)^2}{|\alpha_c|},$$
        where $Z_c$ is the unlearning costs for client $c$ and $\alpha_c$ is the contribution factor of client $c$.
        With the same data contribution for all clients $c$, we have
        $$M_e=\frac{1}{|\mathcal{C}\alpha|}\sum_{c\in\mathcal{C}} (Z_{avg}-Z_c)^2,$$
        where $\alpha$ is the contribution factor for all clients. The $M_e$ score is minimized when $Z_c=Z_{avg}$ for all clients $c$, leading to the reduction to equality.
    \end{proof}
\end{proposition}

\begin{proposition}
    \textbf{Axiom of Continuity}: Fairness metrics $M_e$ is continuous with respect to the unlearning costs and client contribution distance.
    \begin{proof}
        The continuity of $M_e$ is also straightforward. According to the definition of $M_e$
        $$
        M_e\triangleq \frac{1}{|\mathcal{C}|}\sum_{c\in\mathcal{C}} \frac{(Z_{avg}-Z_c)^2}{|\alpha_c|}
        $$
        the continuity of $M_e$ is guaranteed by the continuity of polynomial operations.
    \end{proof}
\end{proposition}

\begin{proposition}
    \textbf{Against the Axiom of Homogeneity}: It is noted that fairness metrics $M_e$ is Fairness measure f(x) is a not homogeneous function of degree 0. That is beacuse the $M_e$ can be understood as the weighted variance of the unlearning costs, which is dependent on the unit of measurement or magnitude of the resource allocation.
\end{proposition}

\begin{proposition}
    \textbf{Axiom of Saturation}: Fairness metrics $M_e$ is saturated, \textit{i.e.}, the value of $M_e$ does not vary with the number of clients in the system.
\end{proposition}
This is evident because the $M_e$ score is the weighted variance of the unlearning costs, which is independent of the number of clients in the system based on central limit theorem. However, it is still noted that the contribution distance, served as the weights, may be influenced. Therefore, comparing the $M_e$ scores between different federated unlearning algorithms under the same FL setting can be more meaningful regarding the fairness of unlearning costs.

\begin{proposition}
    \textbf{Axiom of Partition}: Consider an arbitrary partition of all the clients into two groups. The $M_e$ score for total FL system can be expressed with the $M_{e_1}$ and $M_{e_2}$ scores from two groups, for all possible partitions.
    \begin{proof}
        Let $\mathcal{C}_1$ and $\mathcal{C}_2$ be the two groups of clients satisfying $$\mathcal{C}_1\cup\mathcal{C}_2=\mathcal{C} \text{ and } \mathcal{C}_1\cap\mathcal{C}_2=\emptyset.$$ We record the mean values of both group as $Z_{avg}^{(1)}$ and $Z_{avg}^{(2)}$. The $M_e$ score for the total FL system is
        $$\begin{aligned}
            M_e&=\frac{1}{|\mathcal{C}|}\sum_{c\in\mathcal{C}} (Z_{avg}-Z_c)^2\\
            &= \frac{|\mathcal{C}_1| M_{e_1}+|\mathcal{C}_2| M_{e_2}}{|\mathcal{C}|}\\ &+\frac{|\mathcal{C}_1| (Z_{avg}-Z_{avg}^{(1)})^2+|\mathcal{C}_2| (Z_{avg}-Z_{avg}^{(2)})^2}{|\mathcal{C}|}\\
        \end{aligned}$$
        Therefore, for all possible partitions, the $M_e$ score for total FL system can be expressed with the $M_{e_1}$ and $M_{e_2}$ scores from two groups. The $M_e$ satisfies the axiom of partition.
    \end{proof}
\end{proposition}

\section{Evaluation Details}\label{seca:evadetails}
In this section, we first elaborate on our evaluation settings to provide a comprehensive understanding of the datasets and environment in \S\ref{seca:evasettings}. Next in \S\ref{seca:exptabme}, \S\ref{seca:expfiglc} and \S\ref{seca:expfigcl}, we further demonstrate our experiment results in \S\ref{sec:evaluation} under different hyperparameter or datasets. Finally in \S\ref{seca:mia}, we provide more details of the MIA attacks, cascaded leaving experiments and DPA attacks, which are important issues that can be introduced by unfair FU algorithms.

\subsection{Evaluation Settings}\label{seca:evasettings}

\textbf{Datasets:} We evaluate FedShard on well-known datasets including MNIST, Fashion-MNIST~\cite{fmnist}, CIFAR-10, CIFAR-100, EMNIST~\cite{emnist}, SVHN~\cite{SVHN}, Purchase, and Adult. 

\textbf{Non-IID:} To simulate the non-IID data distribution in FL scenarios, we apply the Label-based Dirichlet Partition (LDA)~\cite{dirichlet}, a method that divides data into multiple subsets based on label distributions using the principles of the Dirichlet process with $\rho$ as the hyperparameter. A lower $\rho$ indicates a higher non-IID degree. In this experiment, we take $\rho=0.1$ to simulate a high non-IID case, $\rho=0.5$ for a moderate non-IID case, and $\rho=0.9$ for an IID case.

\textbf{Number of Clients:} Typically, a federated learning system consists of tens to hundreds of clients. We set the number of clients $K$ to 32, 64, 256 and 512 in our experiments. Large-scale experiments involving over a thousand clients are reserved for future work.

\textbf{Number of Runs for Each Experiment:} To ensure the reliability of our experimental results, we conduct each experiment with multiple independent runs. For experiments in Figure \ref{fig:acc}, Figure \ref{fig:lc}, average unlearning costs ($t$) and $M_e$ value in Table \ref{tab:me}, we evaluate the efficiency three times and report the average result and average $M_e$ value, giving a stable estimate. For experiments in Figure \ref{fig:cl}, Table \ref{tab:mia}, and $M_p$ value in Table \ref{tab:me}, we conduct every attack for five times and report the median results and corresponding $M_p$ values, which robustly handles outliers and ensures that our conclusions are not skewed by rare, anomalous outcomes.

\textbf{Experiment Error:} The errors for efficiency-related experiments (unlearning costs $t$, efficiency fairness $M_e$) are within 5\% of the average value, indicating the stability and reliability of our experimental results. The errors for effectiveness-related experiments (attacks, cascaded leaving simulation, and perfromance fairness $M_p$) are within 7\% of the median value due to the probabilistic nature of these attacks and simulations, which is acceptable in the context of security and privacy evaluations.

\textbf{Environment:} All experiments were developed using Python 3.8 and PyTorch 2.3, and executed on a Linux server with an NVIDIA GeForce RTX3090 GPU and an AMD EPYC 7402 CPU.

\subsection{Rest Experiment Results for Table \ref{tab:me}}\label{seca:exptabme}

In this section, we provide additional results of experiments in \S\ref{sec:exp_efficiency}, with client numbers $K=$32, 64, and 256 in Table~\ref{tab:meK32}, \ref{tab:meK64} and \ref{tab:meK256}, respectively. Experimental results show that FedShard achieves outstanding performance fairness and efficiency fairness no matter the number of clients. Besides, with more clients, FedShard can achieve better unlearning efficiency, \textit{e.g.}, when $K=32$, FedShard unlearns 27\% faster than retraining, while when $K=512$, FedShard unlearns 3.7$\times$ faster than retraining. 

\begin{table*}[ht]
    \centering
    \scriptsize

    \begin{tabular}{|cc|c|c|c|c|c|c|}
        \hline
        && \multicolumn{1}{|c|}{FA} & \multicolumn{1}{c|}{FE} & \multicolumn{1}{c|}{RR} & \multicolumn{1}{c|}{FT}& \multicolumn{1}{c|}{FR} & \multicolumn{1}{c|}{\textbf{FS}}\\
        \hline

        \multirow{3}{*}{MNIST}  &$t$
    &\fbl{9244.6}$\pm$396.1        
    &3099.2$\pm$99.5      
    &2814.6$\pm$101.8    
    &6832.1$\pm$274.3  
    &\ftl{10.13}$\pm$0.39     
    &\ftl{2485.1}$\pm$107.6    \\
    & \tcl{$M_e$}  
    &\tcl{52.44}$\pm$2.11
    &\sfttcl{35.18}$\pm$1.27
    &\tcl{36.87}$\pm$1.21
    &\fbtcl{727.66}$\pm$34.77
    &\tcl{53.17}$\pm$1.95
    &\sfttcl{9.51}$\pm$0.41    \\
    & \tdl{$M_p$} &\sfttdl{5.93}$\pm$0.33      
    &\tdl{9.57}$\pm$0.49      
    &\tdl{9.90}$\pm$0.62
    &\tdl{6.54}$\pm$0.39
   &\fbtdl{14.04}$\pm$0.91
    &\sfttdl{6.08}$\pm$0.22      \\
    \hline

    \multirow{3}{*}{FMNIST}  &$t$   
    &\fbl{9984.4}$\pm$420.3        
    &2915.45$\pm$97.3       
    &\ftl{2214.2}$\pm$66.8      
    &6163.7$\pm$249.6     
    &\ftl{11.83}$\pm$0.39    
    &2508.3$\pm$91.4    \\
    &\tcl{$M_e$}   
   &\tcl{49.31}$\pm$22.75   
    &\tcl{32.64}$\pm$1.36 
     &\tcl{41.83}$\pm$1.26 
      &\fbtcl{447.59}$\pm$21.93
     &\sfttcl{31.39}$\pm$1.14   
    &\sfttcl{11.62}$\pm$0.49   \\
    & \tdl{$M_p$}  
    &\sfttdl{5.85}$\pm$0.30       
    &\tdl{9.38}$\pm$0.62     
    &\tdl{9.82}$\pm$0.58  
    &\tdl{6.46}$\pm$0.41
    &\fbtdl{14.23}$\pm$0.91  
    &\sfttdl{5.95}$\pm$0.20     \\

    \hline

    \multirow{3}{*}{CIFAR-10}  &$t$   
    &\fbl{13459.3}$\pm$566.2
    &4166.8$\pm$147.3
   &4644.7$\pm$140.0
    &8403.9$\pm$391.7
    &\ftl{17.34}$\pm$0.58
    &\ftl{3361.1}$\pm$141.5      \\
    &\tcl{$M_e$}   
    &\tcl{75.64}$\pm$3.14
    &\tcl{41.32}$\pm$1.45
    &\tcl{39.30}$\pm$1.19
    &\fbtcl{1081.73}$\pm$51.71
    &\sfttcl{35.92}$\pm$1.43
    &\sfttcl{10.9}$\pm$0.43     \\
    &\tdl{$M_p$}
    &\sfttdl{6.35}$\pm$0.29
    &\tdl{9.78}$\pm$0.61
    &\tdl{9.23}$\pm$0.55
    &\tdl{6.67}$\pm$0.41
    &\fbtdl{19.06}$\pm$1.21
    &\sfttdl{6.08}$\pm$0.21      \\

    \hline

    \multirow{3}{*}{CIFAR-100}  &$t$   
    &\fbl{29660.2}$\pm$1147.6         
    &7576.1$\pm$269.7       
    &\ftl{6398.2}$\pm$182.3      
    &16822.5 $\pm$740.3    
    &\ftl{18.01}$\pm$0.59    
    &{6859.2$\pm$284.6}      \\
    &\tcl{$M_e$}   
    &\tcl{76.98}$\pm$3.32 
    &\tcl{42.24}$\pm$1.53
    &\tcl{41.58}$\pm$1.26
    &\fbtcl{1103.06}$\pm$52.77    
    &\sfttcl{31.82}$\pm$1.16 
    &\sfttcl{11.23}$\pm$0.44\\
    &\tdl{$M_p$}  
   &\sfttdl{5.68}$\pm$0.31
    &\tdl{11.74}$\pm$0.73
    &\tdl{10.94}$\pm$0.62 
    &\tdl{7.13}$\pm$0.44
    &\fbtdl{22.14}$\pm$1.41  
    &\sfttdl{5.94}$\pm$0.23      \\
    \hline
    \end{tabular}
    \caption{Same setting as Table \ref{tab:me}. This table further shows the experimental errors.}~\label{tab:me2}
\end{table*}

\begin{table}[ht]
    \centering
    \scriptsize
    \setlength{\tabcolsep}{1mm} 
        \begin{tabular}{|cc|c|c|c|c|c|c|}
            \hline
            && \multicolumn{1}{|c|}{FA} & \multicolumn{1}{c|}{FE} & \multicolumn{1}{c|}{RR} & \multicolumn{1}{c|}{FT}& \multicolumn{1}{c|}{FR} & \multicolumn{1}{c|}{\textbf{FS}}\\
            \hline
    
            \multirow{3}{*}{MNIST}
            &$t$&\fbl{521.6}       &160.2     &\ftl{137.6}    &389.3  &\ftl{0.53}   &409.9     \\
            & \tcl{$M_e$}&\sfttcl{17.75}      &\tcl{35.10} &\tcl{28.84}
            &\fbtcl{507.46} &\tcl{62.93} &\sfttcl{4.62}\\
            & \tdl{$M_p$}&\sfttdl{5.60}  &\tdl{12.99}    &\tdl{13.90}   &\tdl{6.14}    &\fbtdl{14.06}   &\sfttdl{5.74}\\
        \hline

        \multirow{3}{*}{FMNIST}&$t$ &\fbl{554.7}      &145.8     &\ftl{136.1}    &367.9   &\ftl{0.56}  &415.0    \\
        &\tcl{$M_e$} &\sfttcl{19.87} 
        &\tcl{39.45} 
        &\tcl{27.85} 
        &\fbtcl{582.42} 
        &\tcl{39.09}  
        &\sfttcl{7.13}     \\
        & \tdl{$M_p$}&\tdl{6.35}     
        &\tdl{12.44}     
        &\tdl{14.90}        
        &\sfttdl{5.89}  
        &\fbtdl{18.80} 
        &\sfttdl{6.30}\\
        \hline
        \multirow{3}{*}{CIFAR-10}&$t$ 
        &\fbl{762.8}    
        &223.1     
        &\ftl{185.7}   
        &525.8 
        &\ftl{0.76}   
        &562.3      \\
        &\tcl{$M_e$} &\sfttcl{20.08} 
        &\tcl{39.66} 
        &\tcl{33.95} 
        &\fbtcl{558.21} 
        &\tcl{46.71}
        &\sfttcl{5.31}  \\
        & \tdl{$M_p$}&\tdl{6.19}&\tdl{11.46}&\tdl{12.15}&\sfttdl{6.17}   &\fbtdl{16.71}&\sfttdl{6.02}     \\
        \hline

        \multirow{3}{*}{CIFAR100}  &$t$   
        &\fbl{1623.0}           
        &405.7     &\ftl{367.7}     &1020.7   &\ftl{0.72} &1147.5      \\
        &\tcl{$M_e$}   &\sfttcl{20.41}     &\tcl{40.37}    &\tcl{34.63}    &\fbtcl{573.43}   &\tcl{53.62}    &\sfttcl{7.13}  \\
        &\tdl{$M_p$}  
        &\sfttdl{5.69}    &\tdl{7.68}    &\tdl{11.47}   &\tdl{6.38}    &\fbtdl{20.79}  &\sfttdl{5.70}     \\
        \hline
    
        \end{tabular}
        \caption{Same metrics and baselines as Table~\ref{tab:me}. This table shows the results when $K=32$ and $\rho=0.1$.}~\label{tab:meK32}
\end{table}

\begin{table}[ht]
    \centering
    \scriptsize

    \setlength{\tabcolsep}{1mm} 
        \begin{tabular}{|cc|c|c|c|c|c|c|}
            \hline
            && \multicolumn{1}{|c|}{FA} & \multicolumn{1}{c|}{FE} & \multicolumn{1}{c|}{RR} & \multicolumn{1}{c|}{FT}& \multicolumn{1}{c|}{FR} & \multicolumn{1}{c|}{\textbf{FS}}\\
            \hline
    
            \multirow{3}{*}{MNIST}  &$t$
        &\fbl{1064.2}         
        &356.3        
        &\ftl{283.6}       
        &748.6     
        &\ftl{2.69}       
        &715.4       \\
        & \tcl{$M_e$}  
         &\tcl{32.37}    
         &\tcl{42.53}  
         &\tcl{35.55}  
         &\fbtcl{774.03}   
          &\sfttcl{29.77}   
       &\sfttcl{6.2}      \\
        & \tdl{$M_p$}    &\sfttdl{6.30}  
         &\tdl{8.55}       
          &\fbtdl{14.39}  
         &\tdl{6.74}  
         &\tdl{11.28}  
         &\sfttdl{5.56}      \\

        \hline

        \multirow{3}{*}{FMNIST}  &$t$   
         &\fbl{1090.4}   
         &326.3  
          &\ftl{285.1}     
       &768.0   
         &\ftl{2.77} 
         &1064.1   \\
        &\tcl{$M_e$}   
        &\sfttcl{31.52}   
         &\tcl{41.28}   
        &\tcl{38.72}   
         &\fbtcl{891.96}   
         &\tcl{46.07} 
         &\sfttcl{6.79}    \\
        & \tdl{$M_p$}  
       &\sfttdl{5.90}       
        &\fbtdl{14.27} 
         &\tdl{12.82}  
         &\tdl{6.95} 
       &\tdl{12.73} 
         &\sfttdl{6.17}     \\

        \hline

        \multirow{3}{*}{CIFAR-10}  &$t$   
         &\fbl{1587.7} 
         &417.8 
        &\ftl{381.6} 
         &1069.3 
          &\ftl{3.89}  
         &1042.2   \\
        &\tcl{$M_e$}   
         &\sfttcl{36.58} 
        &\tcl{49.71}  
        &\tcl{41.83}   
        &\fbtcl{851.73}  
        &\tcl{53.86} 
        &\sfttcl{7.13}  \\
        &\tdl{$M_p$}
        &\sfttdl{6.10}  
       &\tdl{13.36}  
         &\tdl{13.99} 
        &\tdl{6.87}  
        &\fbtdl{21.42} 
         &\sfttdl{5.07} \\

        \hline

        \multirow{3}{*}{CIFAR-100}  &$t$   
        &\fbl{3387.2} 
        &875.5 
        &\ftl{755.6}  
        &2076.4  
         &\ftl{3.64} 
         &2127.3  \\
        &\tcl{$M_e$}   
         &\sfttcl{37.22} 
         &\tcl{51.04} 
         &\tcl{42.65}  
        &\fbtcl{874.96}  
         &\tcl{46.58}   
         &\sfttcl{7.91} \\
        &\tdl{$M_p$}  
        &\tdl{6.16}  
         &\tdl{11.48}  
         &\tdl{6.69}  
       &\sfttdl{6.05} 
        &\fbtdl{19.58}  
         &\sfttdl{5.60}  \\

        \hline
        \end{tabular}
        \caption{Same metrics and baselines as Table~\ref{tab:me}. This table shows the results when $K=64$ and $\rho=0.1$.}~\label{tab:meK64}
\end{table}

\begin{table}[ht]
    \centering
    \scriptsize

    \setlength{\tabcolsep}{1mm} 
        \begin{tabular}{|cc|c|c|c|c|c|c|}
            \hline
            && \multicolumn{1}{|c|}{FA} & \multicolumn{1}{c|}{FE} & \multicolumn{1}{c|}{RR} & \multicolumn{1}{c|}{FT}& \multicolumn{1}{c|}{FR} & \multicolumn{1}{c|}{\textbf{FS}}\\
            \hline
    
            \multirow{3}{*}{MNIST}  &$t$
        &\fbl{4950.3}         
        &1511.7  
         &\ftl{1177.9}   
       &3186.7 
         &\ftl{5.93}     
         &1612.2     \\
        & \tcl{$M_e$}  
         &\tcl{66.94}  
         &\tcl{55.54}  
        &\sfttcl{42.46} 
       &\fbtcl{959.18} 
         &\tcl{66.65}
        &\sfttcl{6.68}     \\
        & \tdl{$M_p$}    &\sfttdl{5.73}   
        &\tdl{9.02}        
         &\fbtdl{14.29} 
        &\tdl{6.61}  
       &\tdl{13.57}  
         &\sfttdl{5.77}      \\
        \hline

        \multirow{3}{*}{FMNIST}  &$t$   
         &\fbl{4813.9}       
        &1180.8      
         &\ftl{1171.1}   
       &3132.1   
         &\ftl{6.60} 
         &1662.0     \\
        &\tcl{$M_e$}   
          &\tcl{72.36}  
         &\tcl{53.17}  
         &\sfttcl{34.15}   
        &\fbtcl{1102.84}  
        &\tcl{45.28}  
         &\sfttcl{7.85}   \\
        & \tdl{$M_p$}  
      &\sfttdl{5.92}      
          &\tdl{11.22}    
        &\tdl{8.14}  
          &\tdl{6.29} 
        &\fbtdl{17.68}  
        &\sfttdl{5.84}  \\

        \hline

        \multirow{3}{*}{CIFAR-10}  &$t$   
         &\fbl{6132.5}  
        &2011.0  
        &\ftl{1575.3} 
          &4227.1 
        &\ftl{9.81}  
        &2294.6      \\
        &\tcl{$M_e$}   
         &\tcl{59.26} 
        &\tcl{65.43}  
          &\tcl{38.21} 
          &\fbtcl{855.10}   
        &\sfttcl{32.84}  
 &\sfttcl{7.91}      \\
        &\tdl{$M_p$}
   &\sfttdl{5.59}  
      &\tdl{6.36} 
       &\fbtdl{16.57} 
         &\tdl{6.40} 
        &\tdl{10.74}  
    &\sfttdl{5.92}      \\

        \hline

        \multirow{3}{*}{CIFAR-100}  &$t$   
        &\fbl{13219.7}       
        &3565.5  
          &\ftl{3019.5} 
        &8249.1  
        &\ftl{8.27} 
        &4730.5     \\
        &\tcl{$M_e$}   
          &\tcl{60.31}   
         &\tcl{66.65}  
          &\tcl{47.96}  
          &\fbtcl{1083.84}   
         &\sfttcl{43.07} 
         &\sfttcl{10.94} \\
        &\tdl{$M_p$}  
         &\sfttdl{5.37} 
       &\tdl{13.77}  
         &\tdl{7.39} 
        &\tdl{6.25}  
        &\fbtdl{24.77}  
        &\sfttdl{5.82}   \\
        \hline

        \end{tabular}
        \caption{Same metrics and baselines as Table~\ref{tab:me}. This table shows the results when $K=256$ and $\rho=0.1$.}~\label{tab:meK256}
\end{table}

Further more, we provide additional results of experimental in \S\ref{sec:exp_efficiency}, with different Dirichlet non-IID levels $\rho=0.5$ and $\rho=0.9$ in Table~\ref{tab:merho05} and \ref{tab:merho09}, respectively. The small $\rho=0.1$ indicates a high non-IID level, and the $\rho=0.9$ implies that clients have almost IID data. Experimental results show that FedShard is robust in efficiency fairness, performance fairness and unlearning efficiency for training data with different non-IID levels.
\begin{table}[ht]
    \centering
    \scriptsize

    \setlength{\tabcolsep}{1mm} 
        \begin{tabular}{|cc|c|c|c|c|c|c|}
            \hline
            && \multicolumn{1}{|c|}{FA} & \multicolumn{1}{c|}{FE} & \multicolumn{1}{c|}{RR} & \multicolumn{1}{c|}{FT}& \multicolumn{1}{c|}{FR} & \multicolumn{1}{c|}{\textbf{FS}}\\
            \hline

            \multirow{3}{*}{MNIST}&$t$ &\fbl{9746.9}     & 3264.5     & 2965.4   & 7197.8 &\ftl{12.08}   & \ftl{2617.9}\\
            &\tcl{$M_e$}&\fttcl{26.48}&\tcl{27.61}&\tcl{26.61}&\fbtcl{359.43}&\tcl{27.82}&\fttcl{10.47}      \\
            & \tdl{$M_p$}&\fttdl{5.63}      &\tdl{15.05}      &\tdl{16.92}&\tdl{6.98}    &\fbtdl{24.70}&\fttdl{6.05}      \\

        \hline

        \multirow{3}{*}{FMNIST}&$t$ &\fbl{10433.7}      & 3046.6     & \ftl{2313.8}   & 6440.1 &\ftl{13.24}   & {2621.2}\\
        &\tcl{$M_e$}  &\tcl{28.94}  &\tcl{26.19}  &\tcl{29.84}  &\fbtcl{387.1} &\fttcl{16.16}  &\fttcl{9.87}\\
        & \tdl{$M_p$}&\fttdl{5.96}&\tdl{16.06}&\tdl{16.64}&\tdl{6.09}&\fbtdl{25.94}&\fttdl{5.84}      \\

        \hline
        \multirow{3}{*}{CIFAR-10}&$t$ &\fbl{14297.5}       & 4425.9     & 4933.1   & 8927.3  &\ftl{16.68}   & \ftl{3570.2}\\
        &\tcl{$M_e$} &\tcl{50.88}  &\tcl{32.58}  &\fttcl{28.71}  &\fbtcl{413.34}&\tcl{30.13}  &\fttcl{12.04}\\
        & \tdl{$M_p$}&\tcl{6.49}    &\tdl{18.15}     &\tdl{19.57}&        \tdl{6.89}&    \fbl{28.90}&\tcl{6.14}      \\
        \hline

        \multirow{3}{*}{CIFAR-100}&$t$ &\fbl{31780.4}       & 8116.5      & \ftl{6854.8}    & 18021.2   &\ftl{17.70}  & {7349.1}\\
        &\tcl{$M_e$}  &\tcl{30.45}  &\tcl{33.13} &\fttcl{30.03}  &\fbtcl{594.8}  &\tcl{30.70} &\fttcl{12.35 }      \\
        & \tdl{$M_p$}&\tdl{6.04}      &\tdl{22.93}      &\tdl{23.28}&        \tcl{5.92}    &\fbl{30.73}&\tcl{6.11}      \\
        \hline
        \end{tabular}
        \caption{Same metrics and baselines as Table~\ref{tab:me}. This table shows the results when $K=512$ and $\rho=0.5$. }~\label{tab:merho05}
\end{table}
\begin{table}[ht]
    \centering
    \scriptsize

    \setlength{\tabcolsep}{1mm} 
        \begin{tabular}{|cc|c|c|c|c|c|c|}
            \hline
            && \multicolumn{1}{|c|}{FA} & \multicolumn{1}{c|}{FE} & \multicolumn{1}{c|}{RR} & \multicolumn{1}{c|}{FT}& \multicolumn{1}{c|}{FR} & \multicolumn{1}{c|}{\textbf{FS}}\\
            \hline

        \multirow{3}{*}{MNIST}  &$t$   
        &\fbl{9211.5}         
        & 3087.9       
          & 2804.5      
         & 6808.2    
          &\ftl{16.79}     
          & \ftl{2476.0}\\
        &\tcl{$M_e$}  
      &\tcl{14.05}  
        &\tcl{17.62}  
        &\sfttcl{12.88}  
       &\fbtcl{194.98}  
       &\tcl{34.61}  
        &\sfttcl{8.96}      \\
        &\tdl{$M_p$}  
       &\sfttdl{5.63}  
       &\tdl{29.42}  
         &\tdl{36.60}  
        &\tdl{6.67}  
       &\fbtdl{40.92}  
         &\sfttdl{6.39}      \\

        \hline

        \multirow{3}{*}{FMNIST}  &$t$   
        &\fbl{9485.2}        
       &2769.7       
       & \ftl{2103.5}      
        &5854.5     
      &\ftl{11.46}    
         &{2382.9}\\
        &\tcl{$M_e$}   
   &\tcl{15.92}   
         &\tcl{16.47}  
         &\tcl{14.27}   
        &\fbtcl{460.03}   
      &\sfttcl{8.86}    
       &\sfttcl{8.45}   \\
        & \tdl{$M_p$}  
     &\sfttdl{6.16}      
    &\tdl{30.63}  
      &\tdl{31.73}  
    &\tdl{6.78}  
       &\fbtdl{35.14}  
      &\sfttdl{5.77}      \\

        \hline
        \multirow{3}{*}{CIFAR-10}  &$t$   
      &\fbl{13039.9}
          &4039.5
         & 4502.1
        & 8146.6
      &\ftl{17.74}     
      & \ftl{3258.3}\\
        &\tcl{$M_e$}   
       &\tcl{16.72}   
        &\tcl{20.79}   
        &\tcl{14.14}   
       &\fbtcl{224.23}   
       &\sfttcl{10.28}    
         &\sfttcl{10.30}     \\

        &\tdl{$M_p$}  
    &\tdl{6.64}
        &\tdl{33.25}
     &\tdl{33.93}  
     &\sfttdl{6.51}  
       &\fbtdl{37.13}  
         &\sfttdl{5.88}      \\
        \hline

        \multirow{3}{*}{CIFAR-100}  &$t$   
        &\fbl{29394.6}
        & 507.9
         &\ftl{6340.5}
         &16668.7
        &\ftl{15.35}    
         &{6797.9}\\

        &\tcl{$M_e$}  
       &\tcl{17.45}   
      &\tcl{21.14}   
        &\tcl{14.52}   
        &\fbtcl{299.82}    
         &\sfttcl{6.91}   
      &\sfttcl{10.57}      \\

        &\tdl{$M_p$}  
        &\tdl{6.63}        
       &\tdl{36.25}          
        &\tdl{36.13}  
      &\sfttdl{6.20}  
        &\fbtdl{42.16}  
       &\sfttdl{5.52}      \\
        \hline
        \end{tabular}
        \caption{Same metrics and baselines as Table~\ref{tab:me}. This table shows the results when $K=512$ and $\rho=0.9$. }~\label{tab:merho09}
\end{table}

We also provide additional results of experimental in \S\ref{sec:exp_efficiency}, with more datasets in Table~\ref{tab:meDS}. The extra datasets include Purchase, Adult, SVHN and EMNIST, which are widely used in FL literature. The results imply that FedShard maintains its fairness and efficiency in different datasets.

\begin{table}[ht]
    \centering
    \scriptsize

    \setlength{\tabcolsep}{1mm} 
        \begin{tabular}{|cc|c|c|c|c|c|c|}
            \hline
            && \multicolumn{1}{|c|}{FA} & \multicolumn{1}{c|}{FE} & \multicolumn{1}{c|}{RR} & \multicolumn{1}{c|}{FT}& \multicolumn{1}{c|}{FR} & \multicolumn{1}{c|}{\textbf{FS}}\\
            \hline

            \multirow{3}{*}{EMNIST}&$t$&\fbl{10252.44}&3236.68
            &3116.81
            &7637.51
            &\ftl{10.56}
            &\ftl{2798.04}\\
            & \tcl{$M_e$}&\tcl{79.63}
            &\tcl{40.51}
            &\fttcl{43.5}
            &\fbtcl{879.08}
            &\tcl{64.6}
            &\fttcl{11.34}\\
            & \tdl{$M_p$}&\fttdl{5.97}
            &\tdl{9.93}
            &\tdl{9.88}
            &\tdl{6.55}
            &\fbtdl{13.94}
            &\fttdl{6.4}\\
            \hline
            \multirow{3}{*}{SVHN}&$t$&\fbl{8399.29}
           &2590.41
            &2543.74
            &6119.78
            &\ftl{8.93}
           &\ftl{2238.55}\\
            & \tcl{$M_e$}
           &\tcl{47.53}
           &\tcl{33.9}
           &\fttcl{35.26}
            &\fbtcl{668.64}
            &\tcl{49.42}
            &\fttcl{8.98}\\
            & \tdl{$M_p$}&\fttdl{6.46}
            &\tdl{9.96}
            &\tdl{11.02}
            &\tdl{6.93}
            &\tdl{15.44}&\fttdl{6.68}\\
            \hline
            \multirow{3}{*}{Adult}&$t$
            &\fbl{531.68}
            &150.33
            &135.31
            &342.29
            &\ftl{2.19}
            &\ftl{86.57}\\
            & \tcl{$M_e$}
            &\tcl{53.76}
            &\tcl{24.55}
            &\fttcl{27.37}
            &\fbtcl{424.86}
            &\tcl{41.28}
            &\fttcl{7.52}\\
            & \tdl{$M_p$}&\fttdl{7.11}
            &\tdl{12.54}
           &\tdl{13.45}
            &\tdl{9.02}
            &\fbtdl{16.89}
            &\fttdl{8.17}\\
            
             \hline
    
            \multirow{3}{*}{Purchase}&$t$&\fbl{46.82}
            &15.28
            &\ftl{5.84}
          &41.48
           &\ftl{0.08}
            &\ftl{8.52}\\
            & \tcl{$M_e$}
            &\tcl{75.19}
           &\fttcl{42.15}
            &\tcl{43.72}
            &\fbtcl{583.63}
            &\tcl{62.27}
            &\fttcl{8.65}\\
            & \tdl{$M_p$}
            &\tdl{10.17}
            &\tdl{16.76}
            &\tdl{18.41}
            &\fttdl{10.14}
            &\fbtdl{24.99}
            &\fttdl{8.06}\\
            
        \hline
        \end{tabular}
        \caption{Same metrics, baselines and setting ($K$=512 and $\rho=0.1$) as Table~\ref{tab:me}. This table shows the results for four extra datasets.}~\label{tab:meDS}
\end{table}

\subsection{Rest Experiment Results for Figure \ref{fig:lc}}\label{seca:expfiglc}
In this section, we also provide the supplementary results of Figure~\ref{fig:lc} under different numbers of clients $K$, different non-IID levels $\rho$ and different datasets. 

\begin{figure}[!h]
    \centering
    \includegraphics[width=0.4\textwidth]{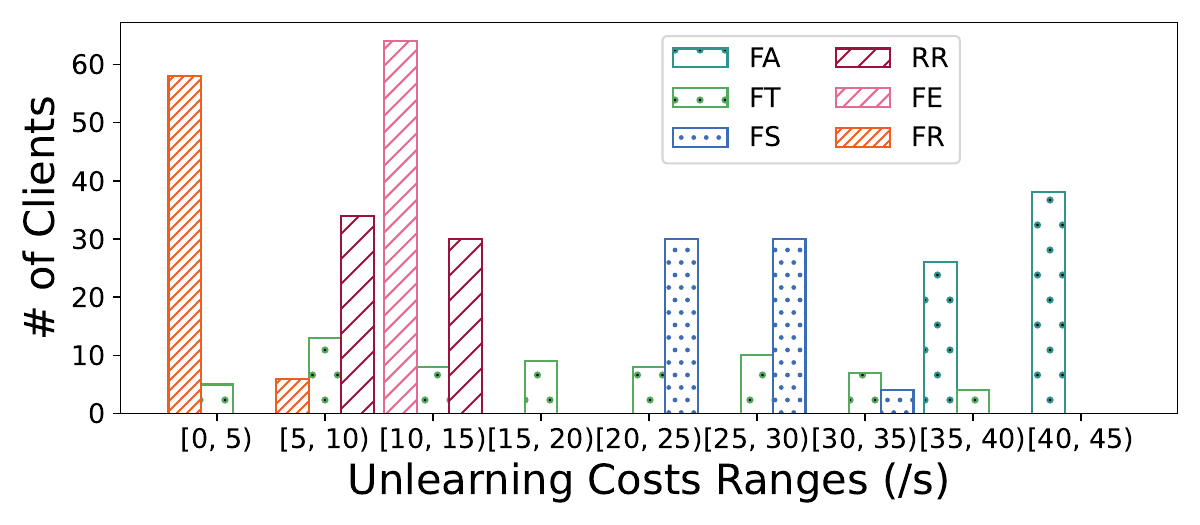}
    \caption{Additional result on CIFAR-10 using 64 clients.}~\label{appfig:13}
\end{figure}

\begin{figure}[!h]
    \centering
    \includegraphics[width=0.4\textwidth]{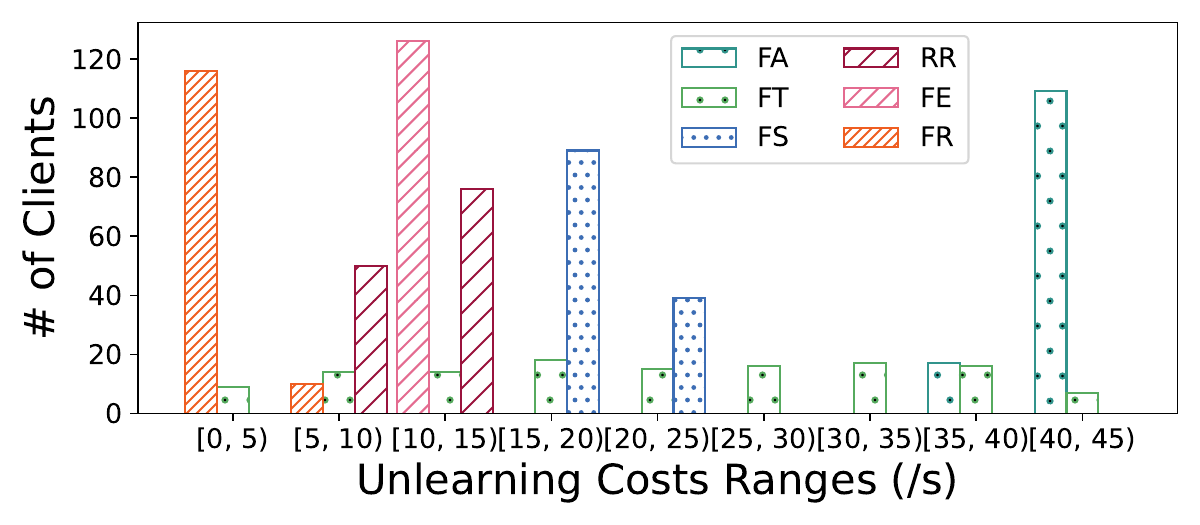}
    \caption{Additional result on CIFAR-10 using 128 clients.}~\label{appfig:14}
\end{figure}

Figures~\ref{appfig:13}, \ref{appfig:14}, and \ref{appfig:14} show the results of Figure~\ref{fig:lc} on CIFAR-10 with 64, 128 and 256 clients, respectively. With more clients, FedShard is much more efficient compared to other retraining methods, and becomes as rapid as the calibration-based methods. Even with fewer clients, FedShard still unlearns faster than other exact retraining methods.
\begin{figure}[!h]
    \centering
    \includegraphics[width=0.4\textwidth]{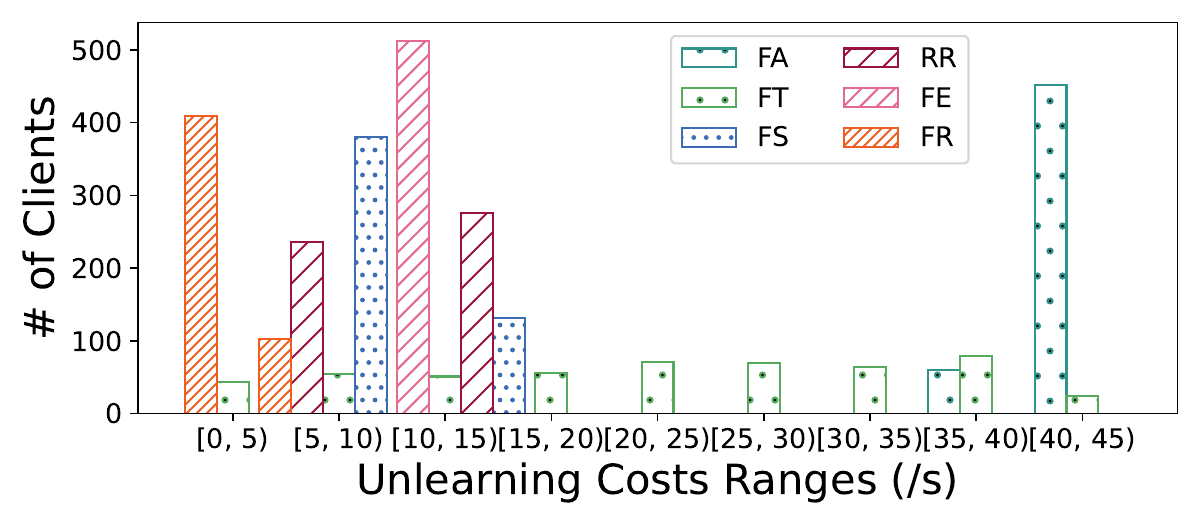}
    \caption{Additional result on CIFAR-10 using 256 clients.}~\label{appfig:15}
\end{figure}

\begin{figure}[!h]
    \centering
    \includegraphics[width=0.4\textwidth]{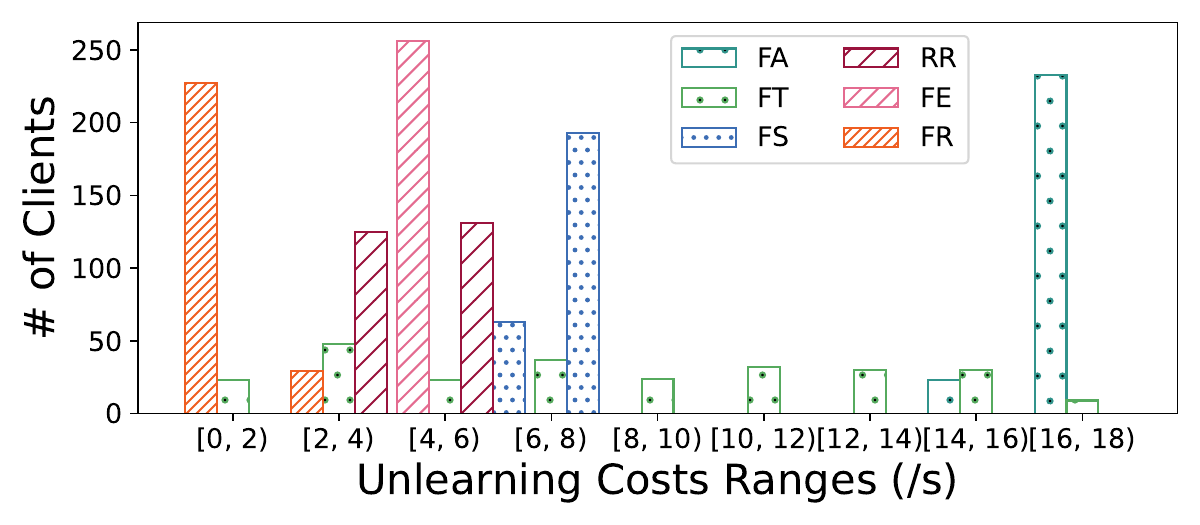}
    \caption{Additional result on MNIST using 256 clients.}~\label{appfig:12}
\end{figure}

\begin{figure}[!h]
    \centering
    \includegraphics[width=0.4\textwidth]{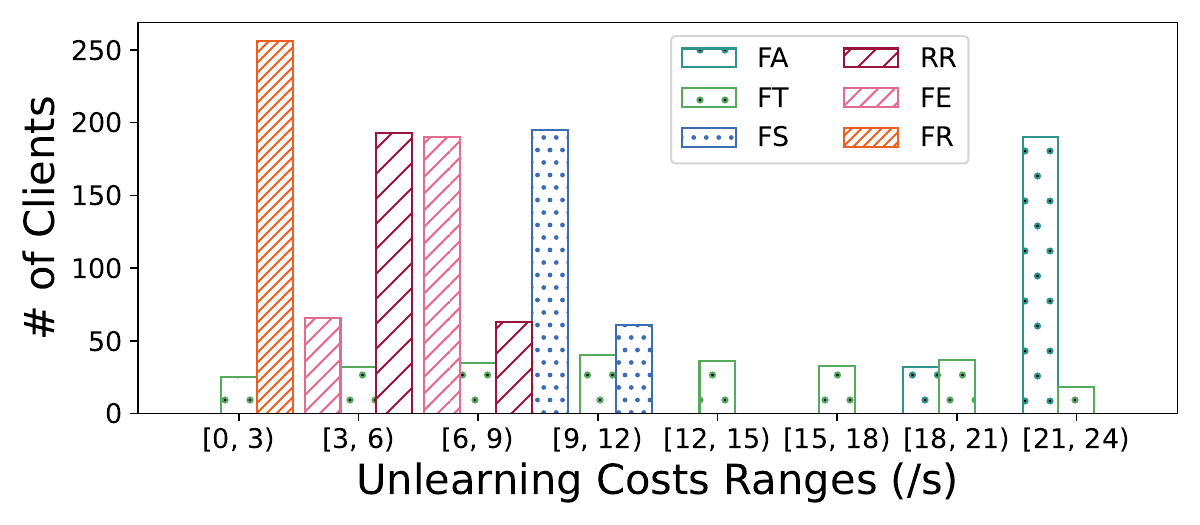}
    \caption{Additional result on FMNIST using 256 clients.}~\label{appfig:18}
\end{figure}

\begin{figure}[!h]
    \centering
    \includegraphics[width=0.4\textwidth]{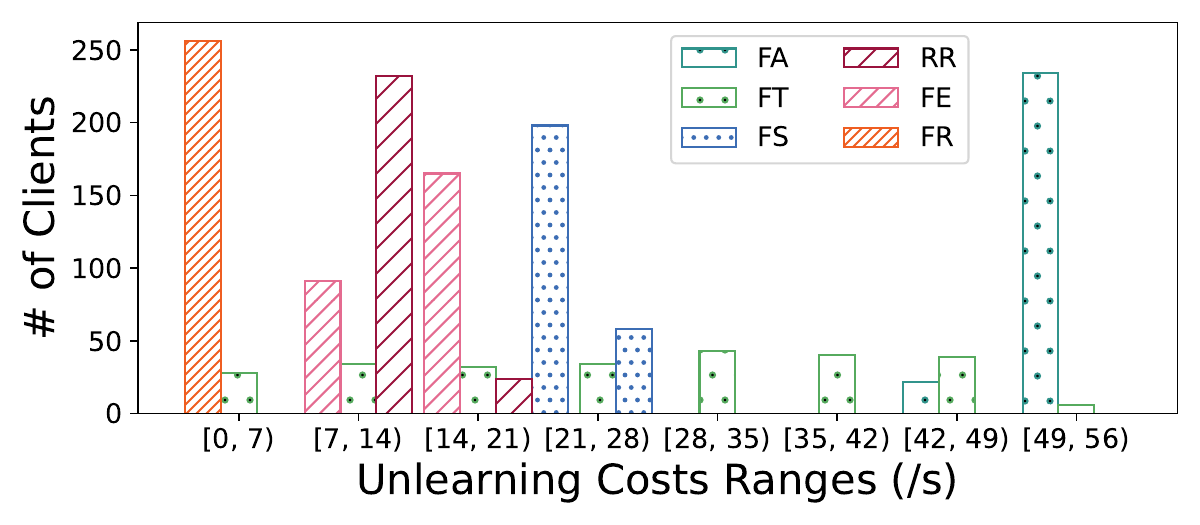}
    \caption{Extra result on CIFAR-100 using 256 clients.}~\label{appfig:9}
\end{figure}

Figures~\ref{appfig:12}, \ref{appfig:18} and \ref{appfig:9} show the results of Figure~\ref{fig:lc} on MNIST, FMNIST and CIFAR-100 with 256 clients, respectively. The results imply that FedShard keeps its efficiency under various datasets.



\subsection{Rest Experiment Results for Figure \ref{fig:cl}}\label{seca:expfigcl}
\begin{figure}[!h]
    \centering
    \includegraphics[width=0.46\textwidth]{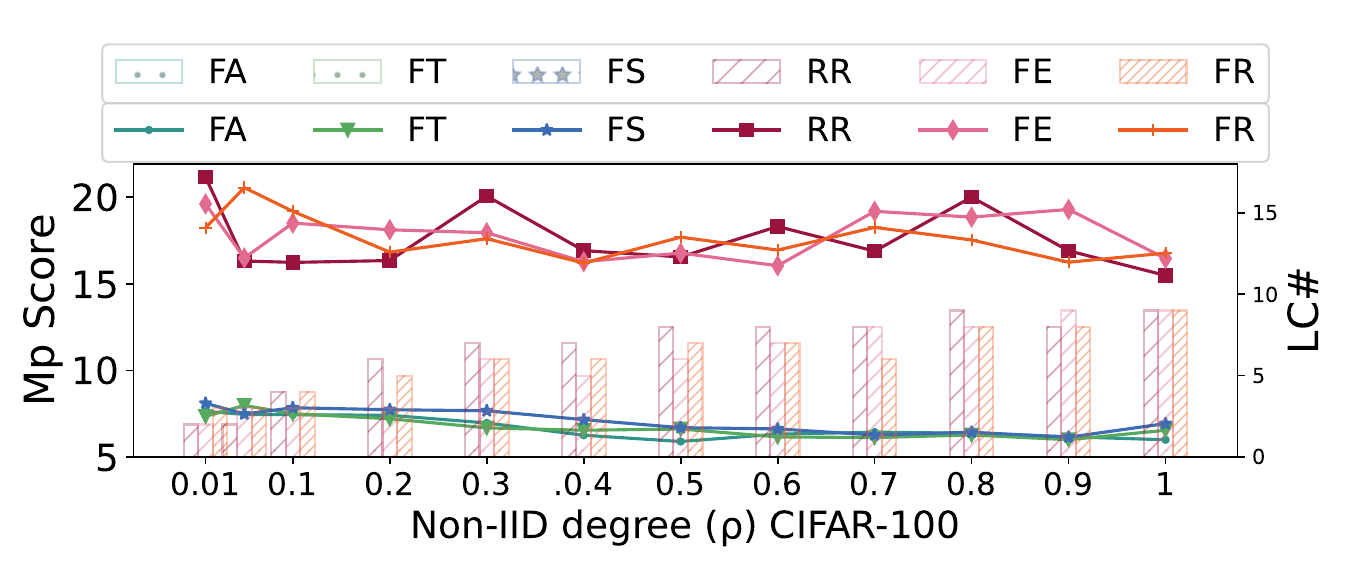}
    \caption{Additional result on CIFAR-100.}~\label{appfig:1}
\end{figure}

\begin{figure}[!h]
    \centering
    \includegraphics[width=0.46\textwidth]{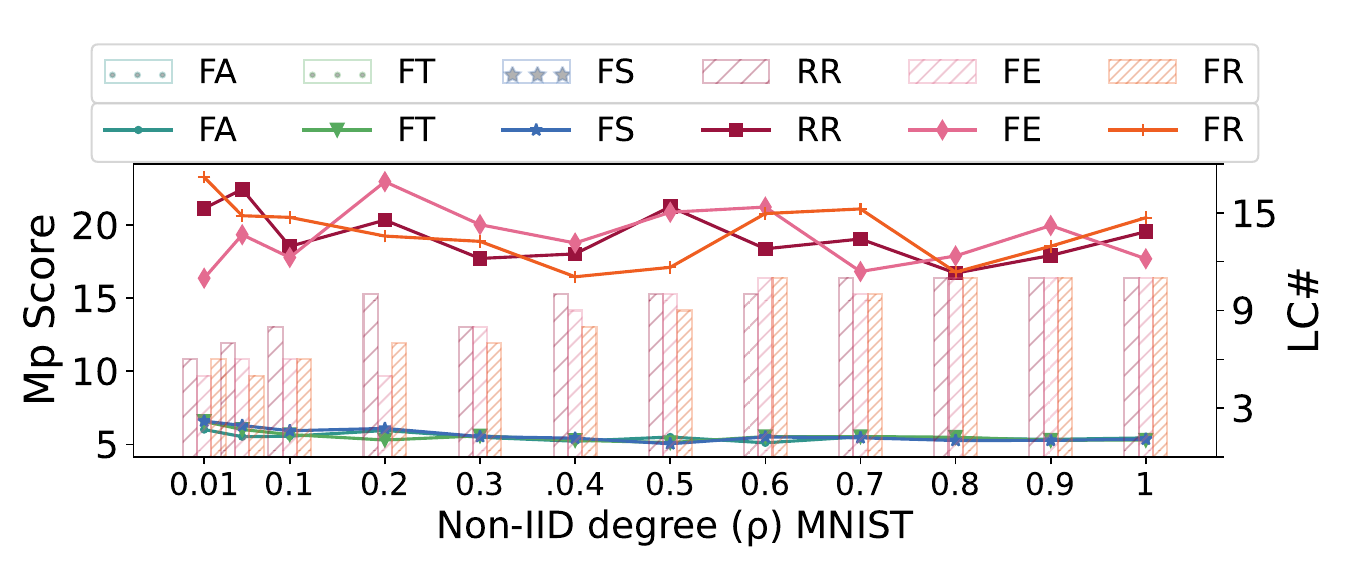}
    \caption{Additional result on FMNIST.}~\label{appfig:2}
\end{figure}

Figures~\ref{appfig:1} and \ref{appfig:2} show the results of Figure~\ref{fig:cl} on CIFAR-100 and FMNIST, respectively. The figures are aligned with our evaluation results, which show that calibration-based methods are faced with the challenge of performance fairness and will cause cascaded leaving. Our FedShard can realize performance fairness and avoid such issues during federated unlearning.

\subsection{MIA Attacks and Cascaded Leaving}~\label{seca:mia}

\textbf{Membership Inference Attacks (MIA)} is a common method used in related works to empirically show the effectiveness of federated unlearning~\cite{fedrec,RR,FATS,federaser,ShardingE,SISA,review1,review2,review3,review4,game}. In MIA, the attacker queries the model with the target sample and the shadow dataset. After that, the attacker compares confidence scores/behavior on member and non-member data to determine if a specific data sample was used to train a model. If one federated unlearning method is effective, it is necessary for it to defend against MIA attacks. Usually, the MIA attacks are evaluated by the F1 score, which is the harmonic mean of precision and recall. The higher the F1 score, the more successful the MIA attacks are. 

\textbf{Cascaded leaving in FL}: Literature~\cite{IncentiveD1, IncentiveD2, IncentiveD3, IncentiveZ, Incentive} shows that strategic clients may leave the FL systems after some clients unlearn their data, which is called cascaded leaving. When one client $c_u$ is leaving though federated unlearning, the payoffs of another client $c_r$ is $W_r^k$=$-\Delta \mathcal{Y} - Z + \lambda_r$ if $c_r$ keeps remaining, and $W_r^l$=$-\mathcal{Y}^{*} + \lambda_r$ if $c_r$ leaves together, where $\Delta \mathcal{Y}$ denotes the performance degradation, $Z$ denotes unlearning cost, $\lambda_r$ represents other payoffs irrelevant to leaving or remaining decisions such as training costs. We do not need to measure the value of $\lambda_r$ when just comparing the difference of payoffs between leaving and remaining, \textit{i.e.}, $W_r^k$ and $W_r^l$. Client $c_r$ will also leave the system when $W_r^k<W_r^l$, \textit{i.e.}, $\Delta Y>Y^*-Z$. If a federated unlearning algorithm is unfair, the large model performance degradation will intensify cascaded leaving.

For the convenience of examining the performance degradation on particular data, we first randomly split the dataset into two partitions by labels, where 48 out of 64 clients own a dataset from one partition and the rest 16 clients own a dataset from another partition. Then we randomly choose 5 clients from the 16 clients to leave. After unlearning, we calculate the payoffs $W_r$ for each remaining client to examine whether it chooses to leave or keeps staying in the system. We further calculate the performance fairness score $M_p$ to evaluate the performance fairness degree. 

\textbf{Data Poisoning Attacks (DPA)}: We propose a new data poisoning attack specific to unfair federated unlearning, named as uf-DPA. If a FU method violates performance fairness, for example, the performance degradation of some clients is much larger than others, then malicious clients (1) first join the FL system and train a model with the target data, (2) then leave the system and request unlearning on the target data. For the remaining clients with similar data to the target data, they will suffer an unfair performance degradation, which indicates a successful unfair-unlearning-triggered DPA.

For the convenience of examining the attack precisions on particular data, we first split the dataset in the same way as in the cascaded leaving experiment. We define the DPA attack precision as the ratio of the number of successfully poisoned clients to the total number of clients. Then we let all 16 clients of the second partition leave, which simulates DPA attackers, and calculate the DPA attack precision for each federated unlearning algorithm. We further calculate $M_p$ scores of each method to validate the effectiveness of our proposed $M_p$ metrics. 

\end{document}